\documentclass{article}
\DeclareUnicodeCharacter{0308}{HERE!HERE!}

\usepackage[nonatbib, preprint]{neurips_2024}

\usepackage[sorting=nyt,style=authoryear,bibencoding=utf8,backend=biber,natbib=true,uniquename=false,uniquelist=false,maxbibnames=99,maxcitenames=1,mincitenames=1]{biblatex}

\PassOptionsToPackage{table,usenames,dvipsnames}{xcolor}

\usepackage{tikz}
\usetikzlibrary{trees}
\usepackage[utf8]{inputenc} 
\usepackage[T1]{fontenc}    
\usepackage{hyperref}       
\usepackage{url}            
\usepackage{booktabs}       
\usepackage{amsfonts}       
\usepackage{nicefrac}       
\usepackage{microtype}      
\usepackage{xcolor}
\usepackage{amsmath}
\usepackage{graphicx}
\usepackage{array}
\usepackage{listings}
\usepackage{caption}
\usepackage{algorithm2e}
\usepackage{pgfplots}
\pgfplotsset{compat=1.18}
\usepackage{subcaption}
\usepackage{booktabs}
\usepackage{multirow}
\usepackage{tabularx}
\usepackage{geometry}
\usepackage{colortbl}
\usepackage{tikz}
\usepackage{booktabs}
\usepackage{lscape}

\definecolor{Gred}{RGB}{219, 50, 54}
\definecolor{ToCgreen}{RGB}{0, 128, 0}

\usepackage[most]{tcolorbox}

\usepackage{tikz}
\usetikzlibrary{positioning, shapes.geometric, arrows.meta, bending, decorations.markings}

\definecolor{lightblue}{rgb}{0.68, 0.85, 0.9}  

\hypersetup{
    colorlinks,
    linkcolor={red!50!black},
    citecolor={blue!50!black},
    urlcolor={blue!80!black},
}

\newcommand{\csLG}{\texttt{cs.LG}}
\newcommand{\astroPH}{\texttt{astro-ph}}

\title{Disentangling Dense Embeddings with\\ Sparse Autoencoders}

\author{%
  Charles O'Neill\thanks{Joint first author contribution.} \\
  The Australian National University\\
  \texttt{charles.oneill@anu.edu.au} \\
  \And
  Christine Ye\footnotemark[1] \\
  Stanford University \\
  \texttt{cye@stanford.edu} \\
  \AND
  Kartheik Iyer \\
  Columbia University \\
  \texttt{kartheikiyer@gmail.com} \\
  \And
  John F. Wu \\
  Space Telescope Science Institute \\
  Johns Hopkins University \\
  \texttt{jfwu@stsci.edu} \\
}

\begin{document}

\definecolor{headercolor}{rgb}{0.2, 0.4, 0.6}
\definecolor{rowcolor1}{rgb}{0.9, 0.9, 0.9}
\definecolor{rowcolor2}{rgb}{1.0, 1.0, 1.0}

\definecolor{low}{rgb}{0.9, 1.0, 0.9}
\definecolor{medium}{rgb}{0.6, 0.9, 0.6}
\definecolor{high}{rgb}{0.3, 0.8, 0.3}

\definecolor{low_green}{rgb}{0.9, 1.0, 0.9}
\definecolor{medium_green}{rgb}{0.6, 0.9, 0.6}
\definecolor{high_green}{rgb}{0.3, 0.8, 0.3}
\definecolor{low_blue}{rgb}{0.9, 0.9, 1.0}
\definecolor{medium_blue}{rgb}{0.6, 0.6, 1.0}
\definecolor{high_blue}{rgb}{0.3, 0.3, 1.0}

\maketitle

\begin{abstract}
Sparse autoencoders (SAEs) have shown promise in extracting interpretable features from complex neural networks. We present one of the first applications of SAEs to dense text embeddings from large language models, demonstrating their effectiveness in disentangling semantic concepts. By training SAEs on embeddings of over 420,000 scientific paper abstracts from computer science and astronomy, we show that the resulting sparse representations maintain semantic fidelity while offering interpretability. We analyse these learned features, exploring their behaviour across different model capacities and introducing a novel method for identifying ``feature families'' that represent related concepts at varying levels of abstraction. To demonstrate the practical utility of our approach, we show how these interpretable features can be used to precisely steer semantic search, allowing for fine-grained control over query semantics. This work bridges the gap between the semantic richness of dense embeddings and the interpretability of sparse representations. We \href{https://huggingface.co/charlieoneill/embedding-saes}{open source} our embeddings, trained sparse autoencoders, and interpreted features, as well as a \href{https://huggingface.co/spaces/charlieoneill/saerch.ai}{web app} for exploring them.
\end{abstract}

\section{Introduction}

The advent of large language models has revolutionised natural language processing, enabling the representation of text in rich semantic spaces \citep{devlin2018bert, brown2020language}. These dense neural vector embeddings capture nuanced semantic relationships, significantly enhancing downstream applications such as information retrieval (IR) and semantic search \citep{reimers2019sentence, gao2022precisezeroshotdenseretrieval, wang2024textembeddingsweaklysupervisedcontrastive}. However, the power of these representations comes at a cost: reduced interpretability and limited user control, presenting significant challenges for fine-tuning and explaining search results \citep{Cao2023STEPGS}.


To address these limitations, recent research has explored methods to disentangle and interpret the information encoded in dense representations \citep{trifonov2018learningevaluatingsparseinterpretable}. Sparse autoencoders (SAEs) have emerged as a promising solution for extracting interpretable features from high-dimensional representations \citep{ng2011sparse, makhzani2013k}. By learning to reconstruct inputs as linear combinations of features in a higher-dimensional sparse basis, SAEs can disentangle complex representations into individually interpretable components. This approach has shown success in analysing and steering generative models \citep{conmysteering2024, thesephistPrismMapping, cunningham2023sparse}, but its application to dense text embeddings remains unexplored.

In this work, we present the first application of SAEs to dense text embeddings derived from large language models. We demonstrate that this approach bridges the gap between the semantic richness of dense embeddings and the interpretability of sparse representations, offering new possibilities for understanding and manipulating textual semantic spaces. In a direct demonstration of their utility for semantic search, we show how SAE features can be used to steer search results, which has been previously been applied to decoder-only transformers and diffusion models for guided generation \citep{elhage2022solu, lesswrongInterpretingSteering}. By causally manipulating features in the SAE's hidden representation of an embedding vector, we can perform precise adjustments of the semantic meaning of the vector upon reconstruction.

Our research makes the following key contributions:

\begin{enumerate}
\item We train SAEs with varying sizes on embeddings from a large corpus of scientific papers, demonstrating their effectiveness in learning interpretable features from dense text representations.
\item We conduct a comprehensive analysis of the learned features, examining their interpretability, behaviour across different model capacities, and semantic properties.
\item We introduce the concept of SAE ``feature families'', clusters of related features that allow for multi-scale semantic analysis and manipulation, and look at how features ``split'' across levels of abstraction.
\item We demonstrate the practical utility of our approach by applying these interpretable features to enhance semantic search, allowing for fine-grained control over query semantics. We develop and open-source this as a tool that implements our SAE-enhanced semantic search system, as well as open-sourcing the underlying SAEs.
\end{enumerate}

This paper is organised as follows: Section \ref{sec:background} provides background on embedding representations and sparse autoencoders. Section \ref{sec:training_and_labelling} details our methodology for training SAEs on text embeddings and analysing the resulting features, and the interpretability metrics of the learned features, as well as scaling behaviour of features across model capacities. Section \ref{sec:feature_families} outlines our identification of feature families and feature splitting. Section \ref{sec:saerch} demonstrates the application of our approach to semantic search. Finally, Section \ref{sec:discussion} discusses the implications of our work and potential future directions.

\begin{figure}
    \centering
    \includegraphics[width=0.95\linewidth, trim={1.85cm 0 0.35cm 0},clip]{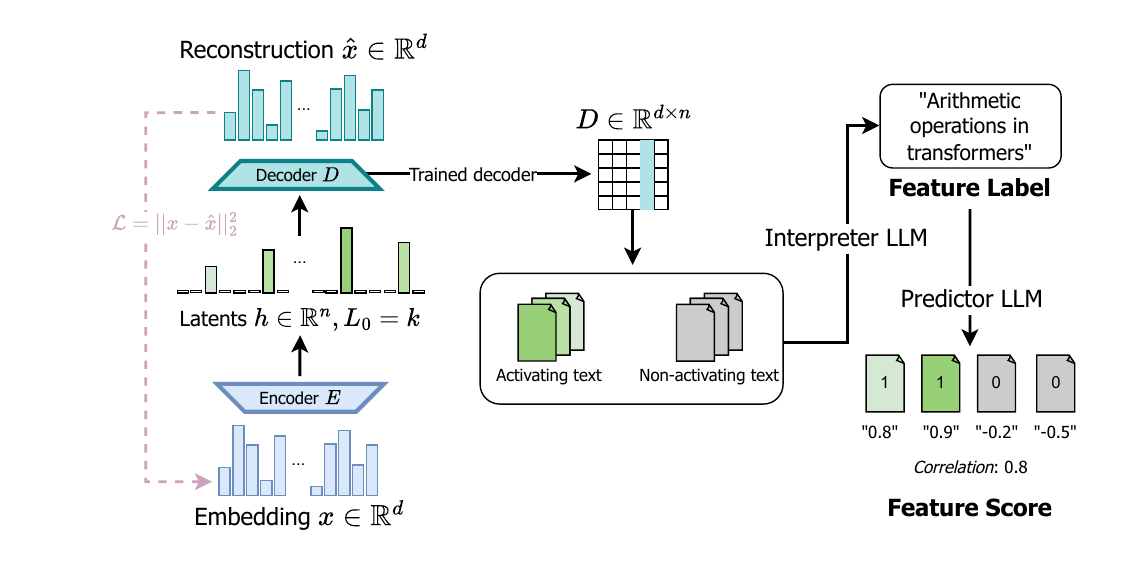}
    \caption{Training and feature labelling process for our sparse autoencoder (SAE). The SAE is trained to minimise reconstruction loss on embeddings from astronomy and computer science paper abstracts. Each feature corresponds to a column in the decoder matrix, representing a direction in embedding space. Feature interpretation involves two steps: (1) An \textit{Interpreter} language model identifies topics present in text that activates each feature but absent in non-activating text. (2) A separate \textit{Predictor} language model assesses feature interpretability by stating its confidence that the feature will activate on unseen text, with confidence correlated with ground truth activations to quantify interpretability.}
    \label{fig:saerch}
\end{figure}

\section{Background and Related work}
\label{sec:background}


Whilst dense embeddings have dramatically improved performance across various NLP tasks, they present significant challenges in terms of interpretability.

\subsection{Embeddings and Representation Learning}

The evolution of word representations in NLP has progressed from simple one-hot encodings to sophisticated dense vector embeddings, each stage offering improvements in semantic expressiveness and contextual understanding. Early distributional semantics models like Latent Semantic Analysis \citep{deerwester1990indexing} paved the way for prediction-based word embeddings such as Word2Vec \citep{mikolov2013efficient} and GloVe \citep{pennington2014glove}, which capture semantic and syntactic relationships. The advent of contextualised word representations, such as ELMo \citep{peters2018deep} and BERT \citep{devlin2018bert}, marked a significant leap by generating dynamic word representations based on surrounding context. Recent developments have extended to sentence and document-level embeddings, with models like Sentence-BERT \citep{reimers2019sentence} proving particularly effective for tasks such as semantic search and information retrieval \citep{gao2021simcse}.

While dense embeddings have dramatically improved the performance of NLP systems, they present challenges in terms of interpretability and fine-grained control. The high-dimensional, continuous nature of these representations makes it difficult to understand or manipulate specific semantic aspects encoded within them \citep{liu2019linguistic}. This opacity can be particularly problematic in applications where explainability or precise semantic control is critical. Semantic search with dense embeddings has largely replaced traditional keyword search \citep{manning2008introduction, baeza1999modern, furnas1987vocabulary, mikolov2013distributed, devlin2018bert, reimers2019sentence}, but faces ongoing challenges, including the ``curse of dense retrieval'' \citep{reimers2022curse} — a phenomenon where the performance of dense retrieval methods degrades more rapidly than sparse methods as index size increases — and lack of interpretability, which hinder fine-tuning of search results \citep{cao2023step, turian2010word}.

\subsection{Sparse autoencoders}

In large language models, the superposition hypothesis suggests that dense neural networks are highly underparameterised, and perform computations involving many more concepts than neurons \citep{elhage2022toymodelssuperposition}. Because these semantic concepts, or \textit{features}, are quite sparse, models compensate encoding multiple features within the same set of neurons. However, this also leads to complex, overlapping representations that are difficult to interpret on a single-neuron basis. Similarly, in embedding spaces, features are not represented monosemantically in individual dimensions. Instead, each feature is typically distributed across multiple dimensions, and conversely, each dimension may contribute to the representation of multiple features. This distributed representation allows embedding models to efficiently encode a large number of features in a relatively low-dimensional space, but it also makes the embeddings challenging to interpret directly.

To address this challenge, sparse autoencoders (SAEs) have emerged as a promising solution. SAEs learn to reconstruct inputs using a sparse set of features in a higher-dimensional space, potentially disentangling superposed features \citep{elhage2022solu, olshausen1997sparse}. By encouraging this disentanglement, SAEs aim to reveal more interpretable and semantically meaningful representations, demonstrating efficacy in uncovering interpretable features in large language model activations \citep{donoho2006compressed, gao2024scaling}. In a well-trained SAE, individual features in the hidden dimension align with the underlying sparse, semantically meaningful features. 


\subsubsection{Architecture and training}

Sparse autoencoders (SAEs) are neural network models designed to learn compact, interpretable representations of high-dimensional data while enforcing sparsity in the hidden layer activations. The architecture of an SAE consists of an encoder network that maps the input to a hidden representation, and a decoder network that reconstructs the input from this representation.

Let $\mathbf{x} \in \mathbb{R}^d$ be an input vector, and $\mathbf{h} \in \mathbb{R}^n$ be the hidden representation, where typically $n \gg d$. The encoder and decoder functions are defined as:
\begin{align}
\text{Encoder}: \quad \mathbf{h} &= f_\theta(\mathbf{x}) = \sigma(W_e\mathbf{x} + \mathbf{b}_e) \\
\text{Decoder}: \quad \hat{\mathbf{x}} &= g_\phi(\mathbf{h}) = W_d\mathbf{h} + \mathbf{b}_d
\end{align}
where $W_e \in \mathbb{R}^{n \times d}$ and $W_d \in \mathbb{R}^{d \times n}$ are the encoding and decoding weight matrices, $\mathbf{b}_e \in \mathbb{R}^k$ and $\mathbf{b}_d \in \mathbb{R}^d$ are bias vectors, and $\sigma(\cdot)$ is a non-linear activation function (e.g., ReLU or sigmoid). The parameters $\theta = \{W_e, \mathbf{b}_e\}$ and $\phi = \{W_d, \mathbf{b}_d\}$ are learned during training.

The training objective of our SAE combines three main components: a reconstruction loss, a sparsity constraint, and an auxiliary loss. The overall loss function is given by:

$$
\mathcal{L}(\theta, \phi) = \frac{1}{d}\|\mathbf{x} - \hat{\mathbf{x}}\|_2^2 + \lambda \mathcal{L}_\text{sparse}(\mathbf{h}) + \alpha \mathcal{L}_\text{aux}(\mathbf{x}, \hat{\mathbf{x}})
$$

where $\lambda > 0$ and $\alpha > 0$ are hyperparameters controlling the trade-off between reconstruction fidelity, sparsity, and the auxiliary loss.

For the sparsity constraint, we use a $k$-sparse constraint: only the $k$ largest activations in $\mathbf{h}$ are retained, while the rest are set to zero \citep{makhzani2013k, gao2024scaling}. This approach avoids issues such as shrinkage, where L1 regularisation can cause feature activations to be systematically lower than their true values, potentially leading to suboptimal representations \textit{shrinkage}, \citep{alignmentforumAddressingFeature, rajamanoharan2024improving}. 

We also use an auxiliary loss, similar to the ``ghost grads'' technique \citep{jermyn2024ghost}, to model the reconstruction error using the top $k_\text{aux}$ dead latents, where we typically set $k_\text{aux} = 2k$. Latents are flagged as dead during training if they have not activated for a predetermined number of tokens (in our case, one full epoch through the training data). Given the reconstruction error of the main model $\mathbf{e} = \mathbf{x} - \hat{\mathbf{x}}$, we define the auxiliary loss as:

$$
\mathcal{L}_\text{aux}(\mathbf{x}, \hat{\mathbf{x}}) = \|\mathbf{e} - \hat{\mathbf{e}}\|_2^2
$$

where $\hat{\mathbf{e}} = W_d \mathbf{z}$ is the reconstruction using the top $k_\text{aux}$ dead latents, and $\mathbf{z}$ is the sparse representation using only these dead latents. This additional loss term helps to revive dead features and improve the overall representational capacity of the model \citep{gao2024scaling}.


\subsubsection{Autointerpretability} \label{sec: background_autointerp}

As SAEs provide a framework for more interpretable representations, researchers have sought ways of automatically interpreting the features that they learn (\textit{autointerpretability}). \citet{gurnee2023finding} introduced sparse probing, where sparse linear models are used to detect the presence of concepts and features in internal model activations through supervised training. Many now use LLMs to directly interpret artifacts of unsupervised models i.e. SAEs. \citet{bills2023language} used GPT-4 to generate and simulate neuron explanations by looking at text on which the neuron activates strongly, while \citet{bricken2023towards} and \citet{templeton2024scaling} applied similar techniques to analyse sparse autoencoder features. \citet{templeton2024scaling} further introduced a specificity analysis to rate explanations by using another LLM to predict activations based on the LLM-generated interpretation, providing a quantification of how interpretable a given neuron or feature actually is. \citet{gao2024scaling} demonstrated that cheaper methods, such as Neuron to Graph \citep{foote2023neuron} with $n$-gram based explanations, allow for a scaleable feature labelling mechanism that does not rely on using more expensive LLMs.

\subsubsection{Structure in SAE features} 
State-of-the-art automated interpretability techniques, as described in Section \ref{sec: background_autointerp}, have resulted in the discovery of a large volume of highly interpretable, monosemantic features in SAEs trained over language models \citep{cunningham2023sparseautoencodershighlyinterpretable, bricken2023towards}. With features being the base unit of interpretability for SAEs, recent work has focused on understanding the geometric structure of features. \citet{bricken2023towards} report \textit{feature splitting} in geometrically close groups of semantically related features, where number of learned features in the cluster increases with model size. They also report the existence of \textit{universal} features which re-occur between independent SAEs and which have highly similar activation patterns. \citet{templeton2024scaling} find feature splitting also occurs in SAEs trained over production-scale models, with larger SAEs also exhibiting \textit{novel} features for concepts that are not represented in smaller SAEs. \cite{makelov2024towards} report \textit{over-splitting} of binary features. \cite{engels2024languagemodelfeatureslinear} find clusters of SAE features that represent inherently multi-dimensional, non-linear subspaces.

\section{Training SAEs and automated labelling}
\label{sec:training_and_labelling}

\subsection{Training and automated interpretability methods}

\textbf{Training:} We train our top-$k$ SAEs on the embeddings of abstracts from papers on arXiv with the \astroPH{} tag (astrophysics, 272,000 papers) and the \csLG{} tag (computer science, 153,000 papers). The embeddings were generated with OpenAI's \texttt{text-embedding-3-small} model.\footnote{\href{https://openai.com/index/new-embedding-models-and-api-updates/}{\texttt{https://openai.com/index/new-embedding-models-and-api-updates/}}} We train our SAEs on these collections of embeddings separately. We normalised the embeddings to zero mean and unit variance before passing them to the SAE as inputs. Our trained SAEs are available for download \href{https://huggingface.co/charlieoneill/embedding-saes}{here}.


\textbf{Hyperparameters:} Notable hyperparameters include the number of active latents $k$, the total number of latents $n$, the number of auxiliary latents $k_\text{aux}$, the learning rate, and the auxiliary loss coefficient $\alpha$. We found learning rate and auxiliary loss coefficient to not have a significant effect on final reconstruction loss; we set the former to 1e-4 and the latter to 1/32. We vary $k$ between 16 and 128, and $n$ between two to nine times the embedding dimension $d_\text{input}$. Whilst we train SAEs with many different combinations of these hyperparameters, we largely focus on what we hereon refer to as SAE16 ($k=16$, $n=2d_\text{input}=3072$), SAE32 ($k=32$, $n=4d_\text{input}=6144$) and SAE64 ($k=64$, $n=6d_\text{input}=9216$). We train each model for approximately 13.2 thousand steps.

\textbf{Automated interpretability:}
Following the training of a Sparse Autoencoder (SAE), it becomes necessary to interpret its features, each corresponding to a column in the learned decoder weight matrix. To facilitate feature interpretation and quantify interpretation confidence, we employ two Large Language Model (LLM) instances: the \textit{Interpreter} and the \textit{Predictor}. The Interpreter is tasked with generating labels for each feature. It is provided with the abstracts that produce the top 5 activations of the feature across the dataset, along with randomly selected abstracts that do not activate the feature. The Interpreter then generates a label for the feature based on this input (for the complete prompt, refer to Appendix \ref{app:automated_interpretability}).
Subsequently, the generated label is passed to the Predictor. The Predictor is presented with three randomly sampled abstracts where the feature was activated and three where it was not. It is then instructed to predict whether a given abstract should activate the feature, expressing its confidence as a score ranging from $-1$ (absolute certainty of non-activation) to $+1$ (absolute certainty of activation).\footnote{We use 3 activating and 3 non-activating abstracts for the Predictor, rather than 5, due to LLM costs. We used \texttt{gpt-4o} as the Interpreter and \texttt{gpt-4o-mini} as the Predictor. Notably, we predict each abstract separately, rather than batching abstracts like \citet{bricken2023towards}.} We measure the Pearson correlation between this confidence and the true activation (binary; +1 or -1). We also measure the F1 score, when framing the confidence as a binary classification (active if confidence is above 0, inactive otherwise).

\textbf{Evaluation metrics:} In order to compare SAEs, we evaluate both their ability to reconstruct the embeddings, as well as the interpretability of the learned features. For the former, we examine the normalised mean squared error (MSE), where we divide MSE by the error when predicting the mean activations. We also report the log density of the activation of features across all papers. We do not report dead latents (those not firing on any abstract) as all models contained zero dead latents at the end of training. We also report the mean activation of features, when their activation is non-zero. To measure interpretability, we use Pearson correlation, as outlined above.

\subsection{SAE Performance}

\begin{figure}
    \centering
    \begin{minipage}{0.49\linewidth}
        \includegraphics[width=\linewidth]{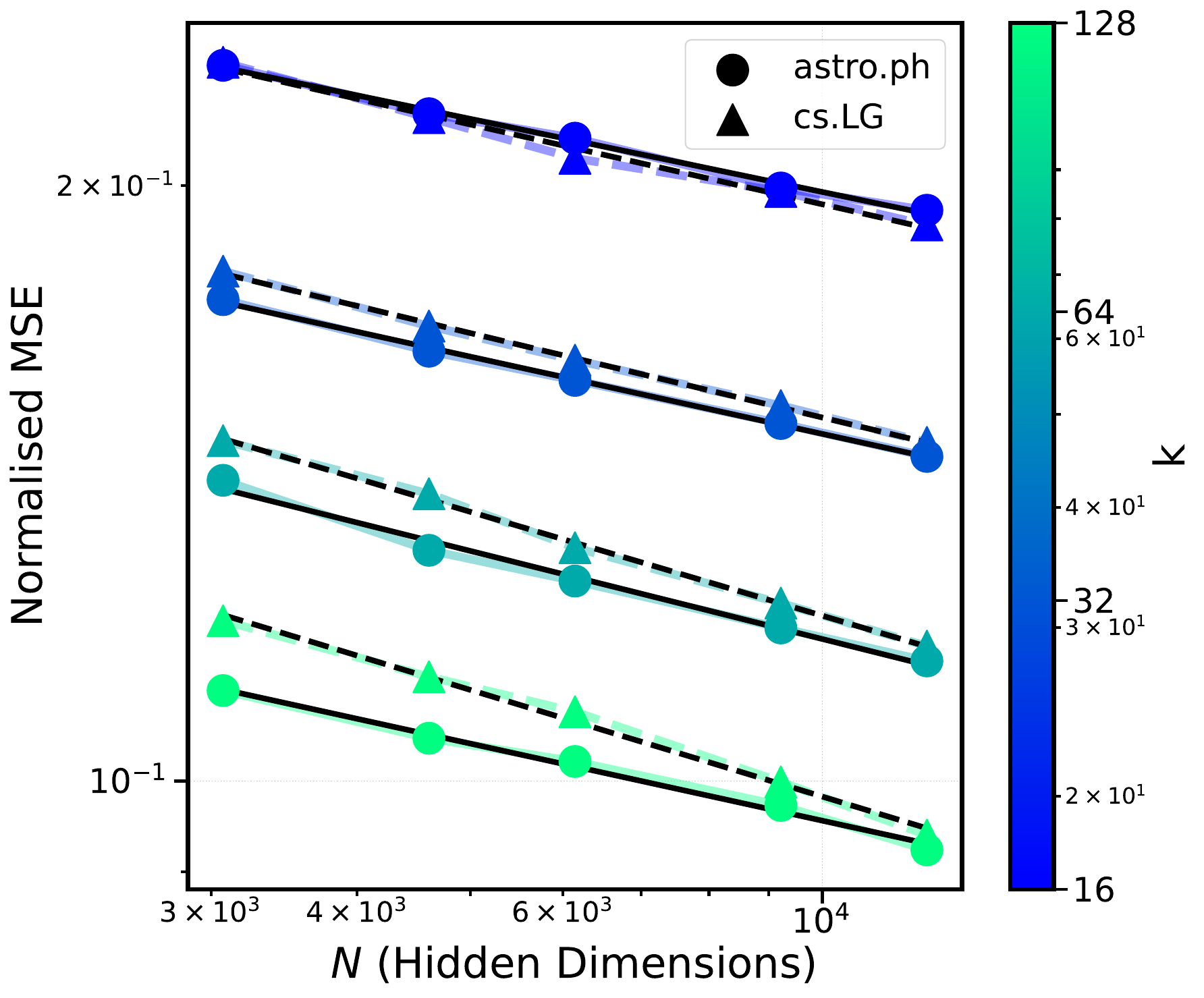}
    \end{minipage}
    \begin{minipage}{0.49\linewidth}
        \includegraphics[width=\linewidth]{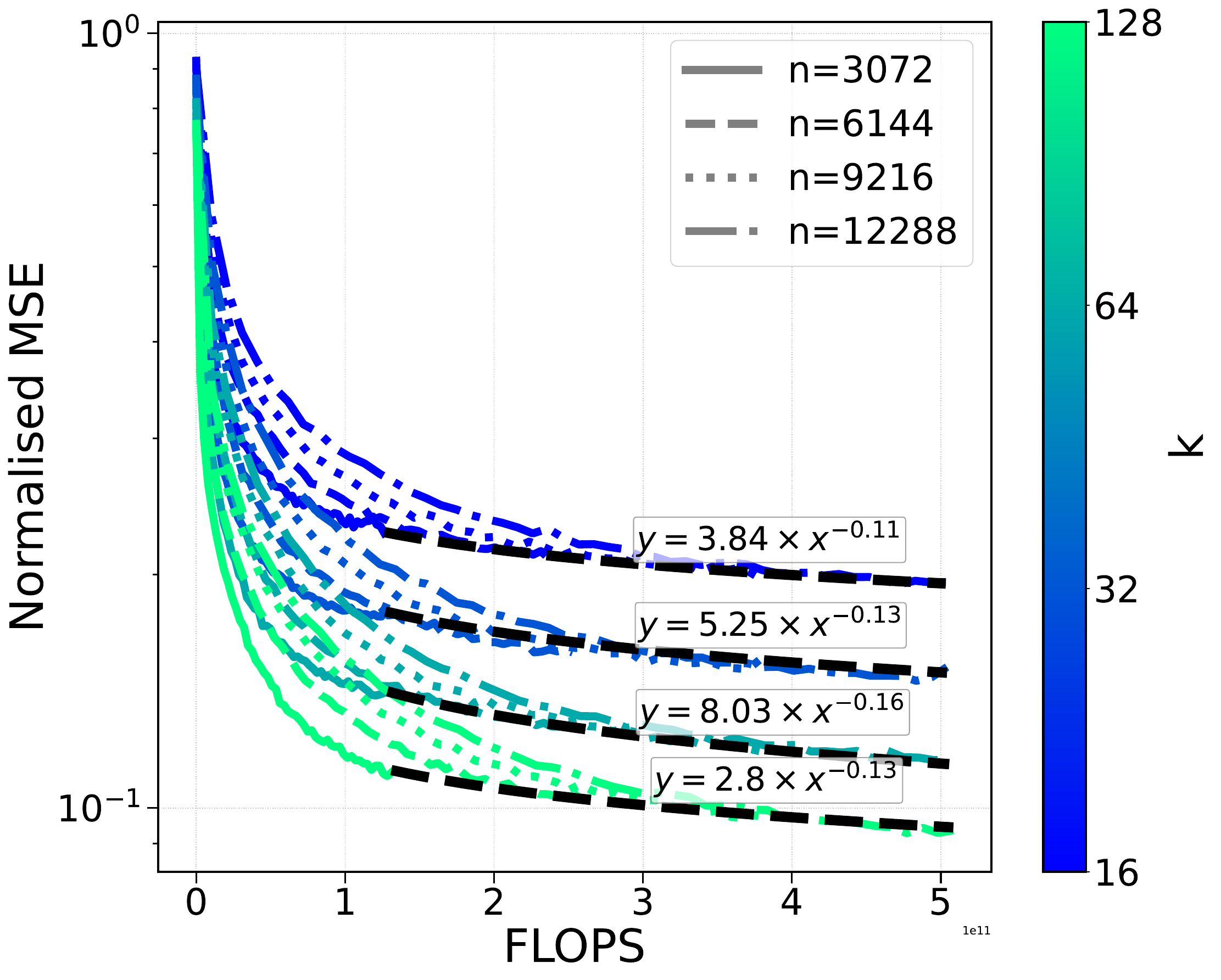}
    \end{minipage}
    \caption{Scaling laws for sparse autoencoder performance. Left: Normalised mean squared error (MSE) as a function of the number of total latents $n$ for different values of active latents $k$. The power-law scaling is evident for each $k$. Right: Reconstruction loss as a function of compute (FLOPs) for different $k$ values, demonstrating the compute-optimal model size scaling.}
    \label{fig:combined_plots}
\end{figure}

We observe precise power-law scalings for sparse autoencoder (SAE) performance as a function of the number of total latents $n$, active latents $k$, and compute $C$ used for training. The normalised mean squared error (MSE) scales as $L(n) = cn^{-\alpha}$ for fixed $k$, where $\alpha$ ranges from 0.12 to 0.18, increasing with $k$, while $c$ generally decreases (Figure \ref{fig:combined_plots}, left panel). The \csLG{} dataset shows slightly higher $\alpha$ values compared to \astroPH{}. For compute scaling, we calculate the number of training FLOPs $C$ at each step for each SAE. We find $L(C) = aC^b$, where $a$ generally increases with $k$ (3.84 for $k=16$ to 8.03 for $k=64$) and $b$ ranges from -0.11 to -0.16, becoming more negative as $k$ increases from 16 to 64 (Figure \ref{fig:combined_plots}, right panel). Both relationships show high accuracy with R-squared values above 0.93. Detailed fits are provided in Appendix \ref{app:training_details}.

\subsection{Interpretability of SAE features}

\begin{figure}
\centering
\begin{minipage}{0.4\textwidth}
    \centering
    \includegraphics[width=\linewidth]{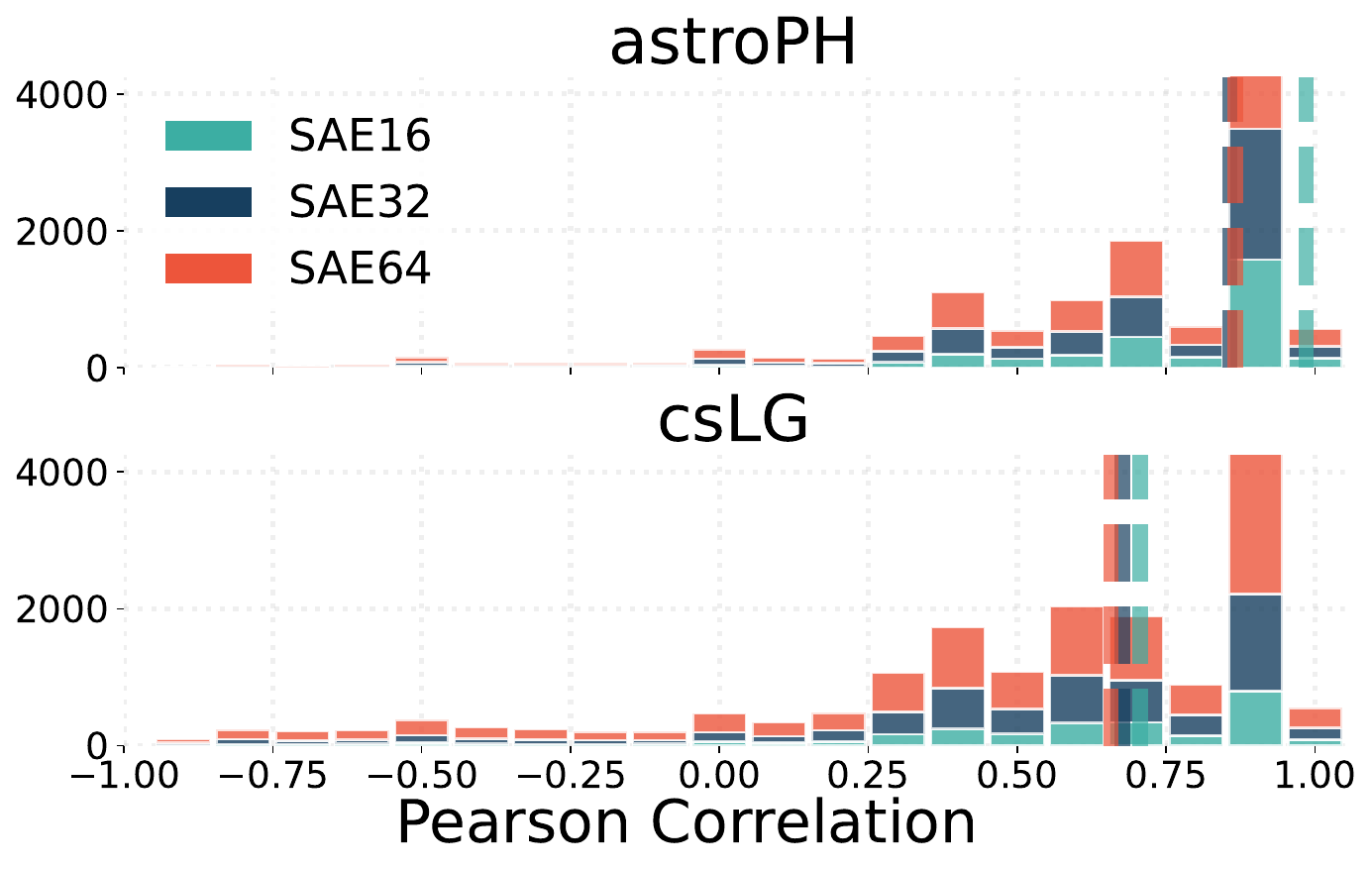}
    \caption{Pearson correlations between the ground-truth and predicted feature activation, using GPT-4o as the \textit{Interpreter} and GPT-4o-mini as the \textit{Predictor}.}
    \label{fig:pearson_correlations}
\end{minipage}%
\hfill
\begin{minipage}{0.56\textwidth}
    \centering
    \includegraphics[width=\linewidth]{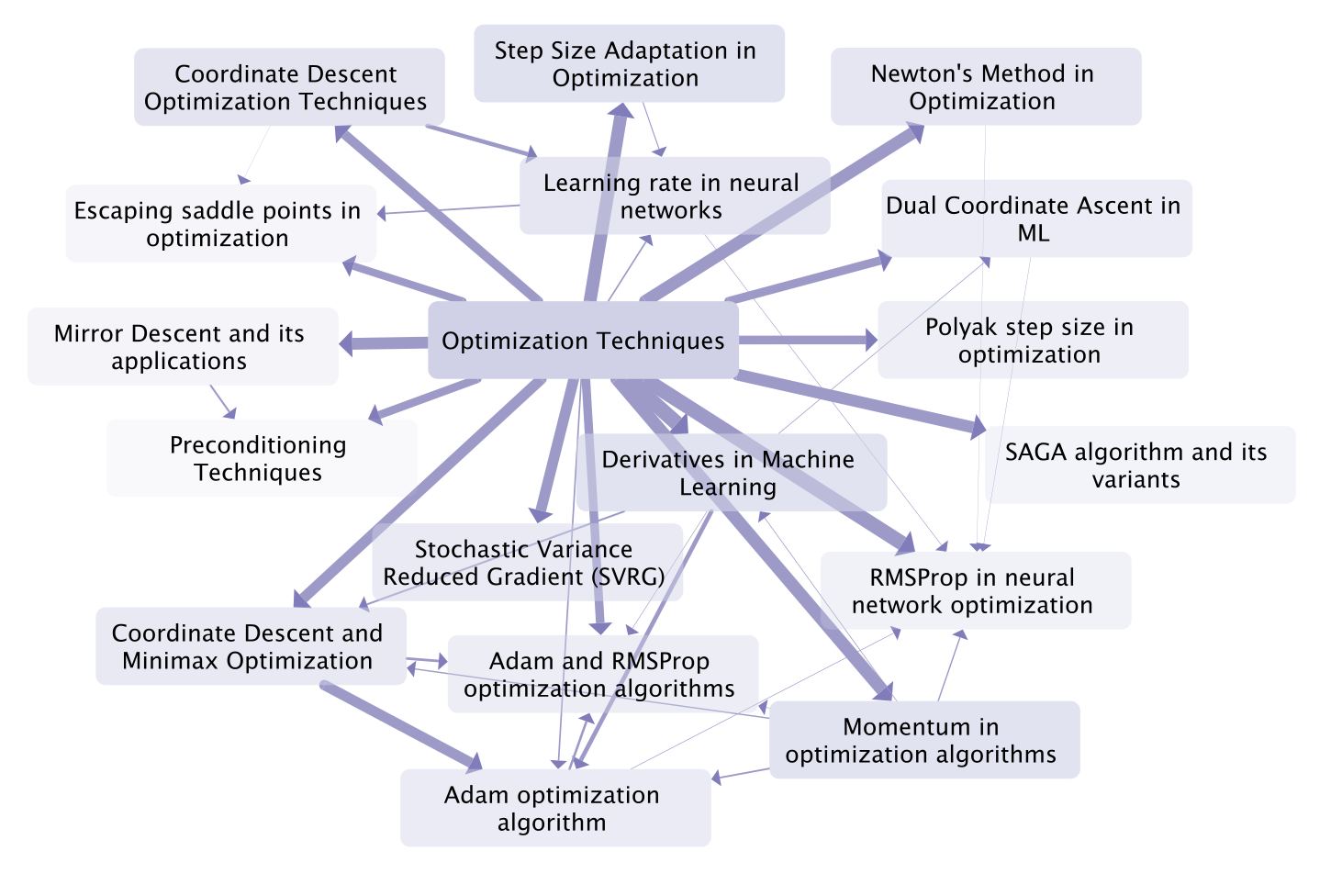}
    \caption{Sample feature family from \texttt{cs.LG}; arrows represent $C_{ij}^{norm} > 0.1$, with size $\propto$  $C_{ij}^{norm}$.}
    \label{fig:featurefamily}
\end{minipage}
\end{figure}

The most direct way to evaluate the interpretability of features is to look at the distribution of automated interpretability scores, discussed above. Specifically: given a feature label from our interpreter model, how well can a predictor model predict the feature's activation on unseen text? We show in Figure \ref{fig:pearson_correlations} that the Pearson correlation between predictor model confidence of a feature firing and the ground-truth firing is quite high, with median correlations ranging from 0.65 to 0.71 for \csLG{} and 0.85 to 0.98 for \astroPH{}. We note that Pearson correlation increases as $k$ and $n$ decrease, likely due to models learning coarser-grained features that are easier for the interpreter to identify.



\section{Constructing feature families through graph-based clustering}
\label{sec:feature_families}
We find that our SAEs trained over arXiv paper embeddings recover a wide range of features. These features cover both scientific concepts, from niche to broad and multi-disciplinary, and also more abstract semantic artifacts, such as humorous writing or critiques of scientific theories. Features and activating examples can be found in Appendix \ref{app:automated_interpretability}.

To analyse how features evolve across different SAE capacities and to identify meaningful groupings of related features, we studied two distinct phenomena: \textit{feature splitting} and \textit{feature families}. \textit{Feature splitting} -- the tendency of features appearing in larger SAEs to ``split'' the direction spanned by a feature from a smaller SAE, and activate on granular sub-topics of the smaller SAE's feature -- has been observed in previous work on sparse autoencoders (e.g. \cite{bricken2023towards}). Examples of feature splitting, as well as features recurring across SAEs, can be found in Figures \ref{fig:splitting} and \ref{fig:splitting2}/\ref{fig:splitting3}. 


In contrast, \textit{feature families} exist within a single SAE, and exhibit a clear hierarchical structure with a dense ``parent'' feature and several sparser ``child'' features; we suggest that the ``parent'' feature encompasses a broader, more abstract concept that is shared among the ``child'' features. An example feature family from \csLG{} can be seen in Figure \ref{fig:featurefamily}. 

\subsection{Feature splitting}

We investigated how features in smaller SAEs relate to features in larger SAEs through a nearest neighbour approach. For each pair of SAEs (i.e. SAE16 and SAE32) with $n_1$ and $n_2$ features respectively, we calculated an $n_1 \times n_2$ similarity matrix $S$ where $S_{ij} = \mathbf{w}_i^T \mathbf{w}_j/\|\mathbf{w}_i\| \|\mathbf{w}_j\|$. Here, $\mathbf{w}_i$ and $\mathbf{w}_j$ are decoder weight vectors for features in the smaller and larger SAE, respectively. For each feature in the larger SAE, we identified the most similar feature in the smaller SAE, allowing us to trace how features potentially ``split'' or become more refined as model capacity increases. 


Our results are shown in Figure \ref{fig:combined_cosine_sim_histograms}. We find that increasing both number of active latents $k$ and the latent dimension $n$ reduces the similarity between nearest neighbours in differently sized SAEs. This agrees with intuition: larger models with more capacity (higher $k$ and $n$) can learn more fine-grained and specialised features, leading to greater differentiation from features in smaller models. 

Qualitatively, matching features from small to large SAEs, we find both \textit{recurrent} features and \textit{novel} features. Recurrent features appear with extremely high $S_{ij}$ and activation similarity across one or more model pairs, and have highly similar auto-generated interpretations, suggesting semantic closeness; these are much more common for lower $k$ (see Figure \ref{fig:splitting}). In contrast, novel features have distinct semantic meaning from their nearest-neighbour match, and activate similarly on some but not all documents; novel features thus \textit{split} the semantic space covered by their nearest-neighbour match from a smaller SAE. However, some novel features share little semantic or activation overlap with their nearest-neighbour feature, as in Fig. \ref{fig:splitting3}, indicating smaller SAEs may not sufficiently cover the feature space; see \ref{section:family} in the Appendix for more details.

\subsection{Feature families}

\paragraph{Feature family identification}
To identify feature families, we developed a graph-based approach using co-activation patterns across the dataset. We consider only highly interpretable features (F1 $\geq 0.8$, Pearson $\geq 0.8$).

We first compute co-occurrence matrix $C$ and activation similarity matrix $D$. For all data points $k$:
\begin{align*}
C_{ij} &= \sum_k A_{ik} A_{jk} & D_{ij} = \sum_k B_{ik} B_{jk}
\end{align*}
where $A_{ik} = 1$ if feature $i$ is active on example $k$ (0 otherwise), and $B_{ik} = \mathbf{h_{k, i}}$ if feature $i$ is active on example $k$ with hidden vector $\mathbf{h_k}$ (0 otherwise). We normalise the co-occurrence matrix by feature activation frequencies and apply a threshold to focus on significant relationships:
\[C_{ij}^{norm} = \frac{C_{ij}}{f_i + \epsilon}\]
where $f_i = \sum_k A_{ik}$ is the activation frequency of feature $i$ and $\epsilon$ is a small constant for numerical stability. We then apply a threshold $\tau$ to obtain $C_{ij}^{thresh}$ (hereafter just $C$). We construct a maximum spanning tree (MST) from $C$, capturing the strongest relationships between features while avoiding cycles. We convert the MST to a directed graph, with edges pointing from higher-density to lower-density features, representing a hierarchy from more general to more specific concepts. We identify feature families via depth-first-search in this directed graph, starting from root nodes (i.e., no incoming edges) and recursively exploring hierarchical sub-families.

We iterate this process, removing parent features after each iteration to re-form the MST and reveal overlapping, finer-grained feature families. We de-duplicate families with high set overlap ($\frac{|F_1 \cap F_2|}{|F_1 \cup F_2|} > 0.6$). In practice, we choose $\tau = 0.1$ and use $n = 3$ iterations.

\paragraph{Feature family interpretability} To evaluate the interpretability of feature families, we analysed their collective properties and the effectiveness of high-level descriptions in capturing their behaviour. For each family, we generated a ``superfeature'' description using GPT-4o, based on the individual feature descriptions within that family. We then uniformly sampled high-activating examples across all activations of child features, and assessed the interpretability of the superfeature using a prediction task, where GPT-4o predicted whether test abstracts would activate the superfeature. We compared these predictions to ground truth activations to compute Pearson correlation and F1 scores. Additionally, we calculated several metrics to characterise the structure and coherence of the feature families. Table \ref{tab:family_interpretability} presents the mean values of these metrics across all families for both the \astroPH{} and \csLG{} datasets.

\begin{figure}
    \centering
    \begin{subfigure}[t]{0.49\textwidth}
    \vskip 0pt
        \centering
        \begin{tikzpicture}
            \node[anchor=south west,inner sep=0] (image) at (0,0) {\includegraphics[width=\textwidth]{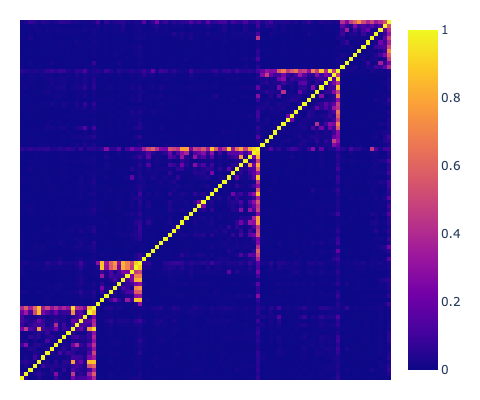}};
            \begin{axis}[
                width=\textwidth*1.15,
                height=\textwidth*1,
                at={(image.south west)},
                anchor=south west,
                xmin=0, xmax=10, ymin=0, ymax=10,
                axis lines=left,
                clip=false,
                ytick = {2.5, 4, 7, 8.5, 10},
                yticklabels = {},
                xtick={2.1, 3.5, 6, 7.5, 9},
                xticklabels={Machine learning, Gamma ray bursts, Black holes, Periodicity detection, Gas dynamics},
                xticklabel style={rotate=45,anchor=east},
                xlabel style={at={(axis description cs:0.5,-0.1)},anchor=north},
                ylabel style={at={(axis description cs:-0.1,0.5)},anchor=south,rotate=90},
            ]
            \end{axis}
        \end{tikzpicture}
        \vspace{0.25cm}
        \caption{}
        \label{fig:subfig1}
    \end{subfigure}
    \hfill
    \begin{subfigure}[t]{0.49\textwidth}
    \vskip 0pt
        \centering
        
        \begin{tikzpicture}
            \node[anchor=south west,inner sep=0] (image) at (0,0) {\includegraphics[width=\textwidth]{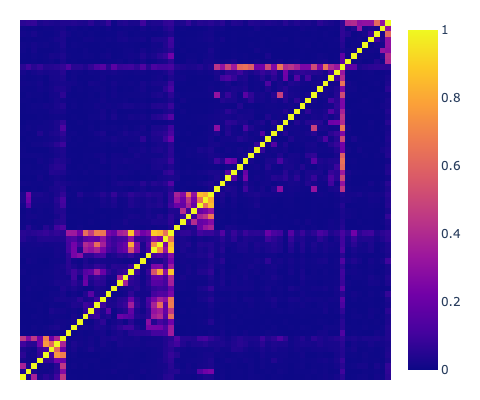}};
            \begin{axis}[
                width=\textwidth*1.05,
                height=\textwidth*1,
                at={(image.south west)},
                anchor=south west,
                xmin=0, xmax=10, ymin=0, ymax=10,
                axis lines=left,
                clip=false,
                ytick = {1.7, 4.5, 5.5, 9, 10},
                yticklabels = {},
                xtick={2.1, 4.5, 6, 8.7, 10},
                xticklabels={Recurrent NNs, Explainability in AI, Recommendation systems, Deep learning for vision, Dimensionality reduction},
                xticklabel style={rotate=45,anchor=east},
                xlabel style={at={(axis description cs:0.5,-0.1)},anchor=north},
                ylabel style={at={(axis description cs:-0.1,0.5)},anchor=south,rotate=90},
            ]
            \end{axis}
        \end{tikzpicture}
        \caption{}
        \label{fig:subfig2}
    \end{subfigure}
    \caption{Co-occurrence matrix $C$ organised by a subset of 5 feature families each. Features in families are ordered by firing density, and the right-most feature is the parent. The un-filled block structure reflects the hierarchical nature of the feature family: all children co-occur with the parent, but few children fire with each other. Visually, this supports our clustering approach.}
    \label{fig:matrix}
\end{figure}

\paragraph{Matrix structure} We conjecture that feature families are equivalent to diagonal blocks in some permutation of the co-occurrence matrix $C$ and activation similarity matrix $D$. If feature families are indeed meaningful clusters in the graph, then in $C$ and $D$ in-block elements should co-activate much more strongly than off-diagonal elements. We also argue that due to the hierarchical nature of feature families, matrix ``blocks'' are highly sparse, since child features all co-occur with the parent feature but rarely co-occur with one another. Subsets of the co-occurrence matrix, permuted by feature family, are shown in \ref{fig:matrix}. 

Motivated by these structures, we compute the parent-child co-occurrence ratio $R(p, \mathcal{C})$ for every family with parent feature $p$ and children $\mathcal{C}$, $\frac{\text{avg}(\sum_{i \in \mathcal{C}} A_{ip})}{\text{avg}(\sum_{i \in \mathcal{C}} \sum_{j \in \mathcal{C}, j \neq i} A_{ij})}$. We also permute the co-occurrence and activation similarity matrices by greedily selecting feature families, and compute the in-block to off-diagonal ratios $C_\text{diag} / C_\text{off}$ and $D_\text{diag} / D_\text{off}$ (excluding the $i = j$ diagonal), capturing the clustering strength of the block diagonal. Statistics are listed in Table \ref{tab:family_interpretability}.

\begin{table}[ht]
\centering
\small 
\begin{tabularx}{\textwidth}{l l l l l l l l l}
\toprule
\textbf{Dataset} & \textbf{($k$, $n$)} & $\mathbf{\text{Size}}$ & $\mathbf{\text{F1}}$ & $\mathbf{\text{Pearson}}$ & $\overline{R(p, \mathcal{C})}$ & $C_\text{diag} / C_\text{off}$ & $D_\text{diag} / D_\text{off}$ & \textbf{$f_{\text{inc}}$} \\
\midrule
\astroPH{} & (16, 3072) & 6 & 0.86 & 0.76 & 10.99 & 5.13 & 5.47 & 0.36 \\
 & (32, 6144) & 6 & 0.86 & 0.73 & 11.75 & 4.72 & 5.87 & 0.31 \\
  & (64, 9216) & 7 & 0.80 & 0.7 & 6.87 & 2.0 & 3.05 & 0.24 \\
\csLG{} & (16, 3072) & 5 & 0.73 & 0.6 & 2.44 & 8.35 & 0.89 & 0.23 \\
& (32, 6144) & 5 & 0.73 & 0.59 & 3.5 & 7.33 & 1.07 & 0.30 \\
& (64, 9216) & 7 & 0.80 & 0.71 & 1.22 & 1.78 & 2.57 & 0.41 \\
\bottomrule
\end{tabularx}
\caption{Interpretability and structure metrics for feature families from \astroPH{} and \csLG{}; we report medians unless otherwise noted. $f_{inc}$ refers to the fraction of features belonging to a clean (F1 $\geq 0.8$, Pearson $\geq 0.8$) feature family.}
\label{tab:family_interpretability}
\end{table}

\color{black}

\section{Evaluating effectiveness of search interventions}
\label{sec:saerch}

\subsection{Intervening on embeddings with SAE features}

As an implementation detail, we note that intervening on a feature by up- or down-weighting its hidden representation and then decoding is equivalent to directly adding the scaled feature vector to the final embedding. We explore an alternative process in Appendix \ref{app:iterative_encoding} where we iteratively optimise the encoded decoded latents to minimise the difference between the desired feature activations and the actual activations.

\subsection{Experiments}

We incorporate SAE-based embedding interventions into a literature retrieval system for \csLG{} and \astroPH{}. To assess the effectiveness of SAE feature intervention on semantic search, we evaluate the \textit{specificity} and \textit{interpretability} of feature-centric query modifications. We select random samples ($N = 50$ each) real literature retrieval queries relevant to machine learning and astronomy, which are answerable with information in papers from \csLG{} and \astroPH{}. For each query, we return the top $k = 10$ most relevant papers using embedding cosine similarity, making up the original retrieval results $\mathcal{R}$. We then select a random feature $i$ in the top-$k$ from the query's hidden representation $\mathbf{h_q}$, and another orthogonal feature $j$ that has no overlap with the top-$k$; we limit our selection only to features that are highly interpretable (F1 $> 0.9$, Pearson $> 0.9$). Given these features, we create a modified query embedding with $\mathbf{h'_{q,i}} = \lambda_{-}$ and $\mathbf{h'_{q,j}} = \lambda_{+}$, letting $\lambda_{-} = 0$ and sampling $\lambda_{+} \in [0, 5]$. This effectively ``down-weights'' and ``up-weights'' the importance of $i$ and $j$, respectively, in the modified query, which is used to generate new retrieval results $\mathcal{R'}$.

To the effect of up-weighting and down-weighting query modifications on the end retrieval results, we provide both $\mathcal{R}$ and $\mathcal{R'}$ to an external LLM instance. The external LLM then compares $\mathcal{R}$ and $\mathcal{R'}$ and determines which features, out of a multiple-choice subset of 5 options, have been up-weighted or down-weighted; we use this to compute the intervention accuracy, which measures the precision and efficacy of causal query interventions. As a baseline, we compare our SAE-based method against traditional query rewriting, by using another LLM instance to re-write the original query such that it up-weights $j$ and down-weights $i$ entirely using natural language.

\begin{figure}
    \centering
    \begin{subfigure}[t]{0.48\textwidth}
        \includegraphics[width=\linewidth]{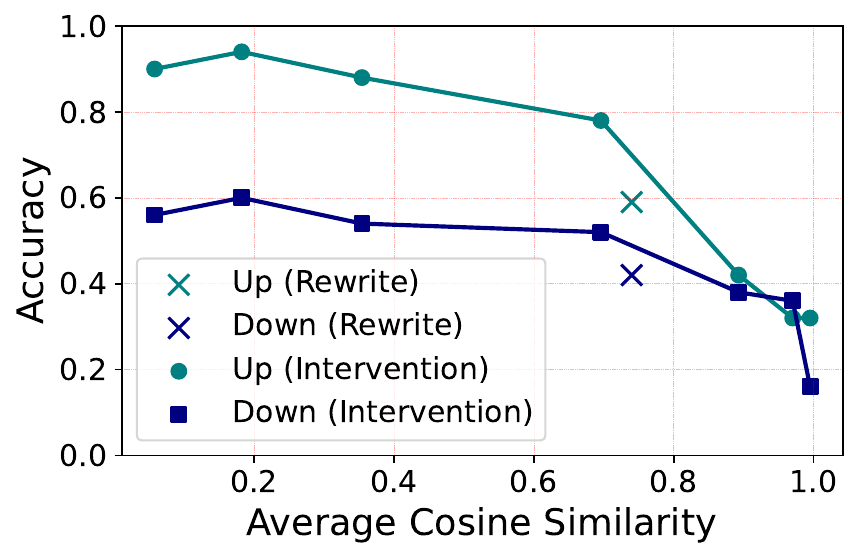}

    \end{subfigure}
    \hfill
    \begin{subfigure}[t]{0.48\textwidth}
        \includegraphics[width=\linewidth]{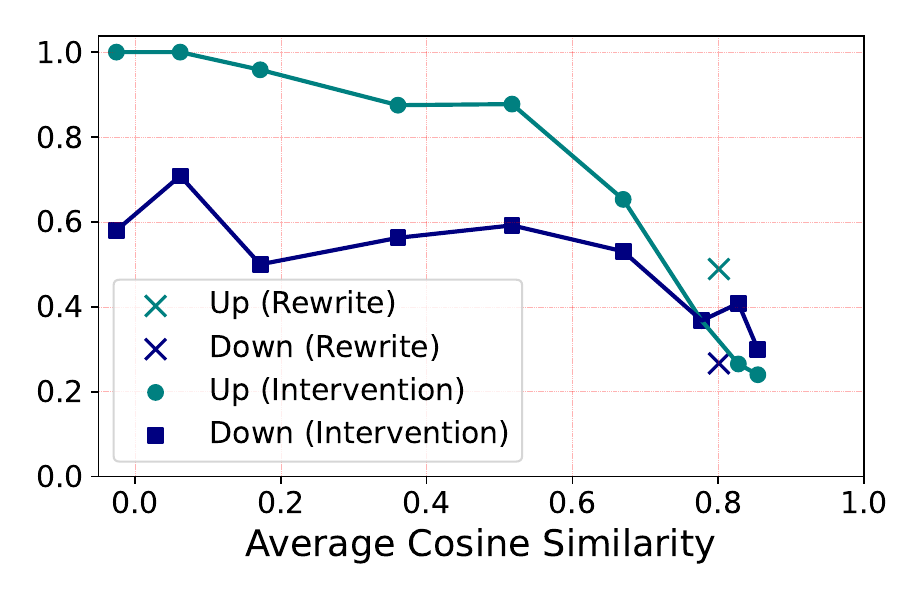}
    \end{subfigure}
    \caption{Relationship between intervention accuracy and query fidelity for SAE-based embedding interventions versus traditional query rewriting in literature retrieval for \csLG{} and \astroPH{} domains. Intervention accuracy measures the precision of causal query modifications, while query fidelity is quantified by cosine similarity between original and modified query embeddings.}
    \label{fig:experiment}
\end{figure}

Our results are shown in Figure \ref{fig:experiment}. We find that SAE feature interventions consistently outperform traditional query rewriting across various levels of query fidelity. This Pareto improvement demonstrates that our method can achieve higher intervention accuracy while maintaining greater similarity to the original query. For instance, at a cosine similarity of 0.75, SAE interventions achieve approximately 20\% higher accuracy compared to query rewriting. 


\color{black}

We also experiment with intervening on feature families, sampling highly interpretable families containing features in the query top-$k$. We uniformly adjust weights for all features in the family, including the parent, using the auto-generated family interpretation as the multiple-choice option. Results show that feature family interventions achieve accuracy comparable to individual features, but only down-weighting interventions outperform query re-writing. This may be because feature families can comprehensively down-weight related concepts, while up-weighting a general concept doesn't necessarily require activating all granular child features. Notably, lower cosine similarity isn't inherently undesirable, as changing the query will naturally reduce similarity. These experiments simply demonstrate that SAE features provide a way to parameterise semantic aspects of a query, allowing controlled modifications of its meaning.


\section{Discussion}
\label{sec:discussion}

In this work, we have presented the first application of sparse autoencoders (SAEs) to dense text embeddings derived from large language models. By training SAEs on embeddings of scientific paper abstracts, we have demonstrated their effectiveness in disentangling semantic concepts while maintaining semantic fidelity. We introduced the concept of ``feature families'' in SAEs, which allow for multi-scale semantic analysis and manipulation. Furthermore, we showcased the practical utility of our approach by applying these interpretable features to enable fine-grained control over query semantics. This aligns with recent work on controlled text generation, where \citet{thesephistPrismMapping} proposed using SAE-based interventions on embeddings to precisely alter generated text in specific semantic directions.

Importantly, our approach offers a novel solution to a growing challenge in scientific literature exploration. With the exponential proliferation of research papers, traditional methods of browsing and discovering relevant literature are becoming increasingly ineffective \citep{tsang2016changing}. Our SAE-based method provides a new way to navigate and find pertinent papers, especially when researchers are not entirely certain about what they're looking for—a key function that libraries used to serve \citep{dahl2021evolving}. By allowing users to explore and manipulate interpretable semantic features, our system enables a more intuitive search process. This capability is particularly valuable in interdisciplinary research or when exploring emerging fields, where relevant work may not be easily discoverable through conventional keyword searches or citation networks \citep{sharma2022research, thomsett2016academic}.

By providing a proof-of-concept for extracting interpretable features from dense embeddings, our work opens several promising research directions and applications across various NLP tasks. In text classification, our method could offer fine-grained insights into model decision boundaries with global features. For machine translation, it could enable targeted semantic manipulations, potentially addressing issues like gender bias in translations \citep{stanovsky2019evaluating}. Scaling this approach to larger and more diverse datasets could reveal insights into how feature families and interpretability evolve with data size and domain breadth. While our current SAEs are trained on narrow scientific domains, extending this to the entirety of arXiv or even internet-scale text corpora could yield general-purpose SAEs with exceptionally rich feature spaces, providing insights into the structure of human knowledge as represented in large-scale text embeddings \citep{thompson2020analysis}. Beyond these applications, our work contributes to the broader goal of making language models more transparent and controllable, which is crucial for building trust in AI systems as they become more integrated into critical decision-making processes \citep{doshivelez2017rigorousscienceinterpretablemachine}.

On an evaluation note, we'd like to see how our reconstructed embeddings fare on a standard semantic embedding benchmark such as MTEB \citep{muennighoff2022mteb} in comparison to the original underlying embeddings. We'd also like to be able to conduct an evaluation of SAE features against some proxy of ground-truth features, much like \citet{makelov2024towards} propose. For instance, the Unified Astronomy Thesaurus \citep{frey2018unified} could provide a basis for evaluating individual feature overlap with astronomy concepts, and even family features as groupings of these individual concepts.

In regard to scientific discovery, our method offers a powerful tool for analysing the evolution of scientific fields over time. By examining statistics of SAE features—such as clustering patterns, co-occurrences, and temporal trends—researchers could gain novel insights into how scientific domains have changed and interacted. This approach could significantly enhance existing efforts to map the landscape of scientific research, which have relied primarily on citation networks and keyword analysis \citep{boyack2005mapping}. For instance, in astronomy, \citet{sun2024knowledgegraphastronomicalresearch} have used knowledge graphs to track field evolution; our SAE-based method could provide a more nuanced and interpretable view of conceptual shifts within the field. Furthermore, the interpretability offered by our SAEs could facilitate interdisciplinary research by providing a common semantic framework for comparing concepts across different scientific domains \citep{uzzi2013atypical}.

\subsection{Limitations}

Our work focused on relatively small datasets from specific scientific domains. Although this specificity allowed us to demonstrate the effectiveness of our approach in targeted areas, future work should investigate how well these methods generalise to larger, more diverse datasets. Additionally, our automated interpretability process, while effective, does not utilise the full spectrum of activations, potentially missing nuanced patterns in feature behaviour. 

The computational requirements for training SAEs on large embedding datasets also present scalability challenges that need to be addressed for wider adoption of this approach. Our SAEs are quite small in comparison to more general language model SAEs. This proved adequate given the narrow domains we trained on, but SAEs for general text embeddings would need to be scaled up by at least 2-3 the total number of latents.

\textbf{Acknowledgments.} Part of this work was done at the 2024 Jelinek Memorial Summer Workshop on Speech and Language Technologies and was supported with discretionary funds from Johns Hopkins University and from the EU Horizons 2020 program’s Marie Sklodowska-Curie Grant No 101007666 (ESPERANTO). Advanced Research Computing at Hopkins provided cloud computing to support the research. We are also grateful for the support from OpenAI through the OpenAI Researcher Access Program. Christine Ye is supported by the Regeneron Science Talent Search. Charles O'Neill is supported by a Tuckwell Scholarship at the Australian National University.  Support for KI was provided by NASA through the NASA Hubble Fellowship grant HST-HF2-51508 awarded by the Space Telescope Science Institute, which is operated by the Association of Universities for Research in Astronomy, Inc., for NASA, under contract NAS5-26555. 

\printbibliography

\newpage

\tableofcontents



\appendix

\section{Training details}
\label{app:training_details}

\subsection{Training setup}
Our sparse autoencoder (SAE) implementation incorporates several recent advancements in the field. Following \citet{bricken2023towards}, we initialise the bias $b_{pre}$ using the geometric median of a data point sample and set encoder directions parallel to decoder directions. Decoder latent directions are normalised to unit length at initialisation and after each training step. For our top-$k$ models, based on \citet{gao2024scaling}, we set initial encoder magnitudes to match input vector magnitudes, though our analyses indicate minimal impact from this choice.

\begin{figure}
    \centering
    \includegraphics[width=0.85\linewidth]{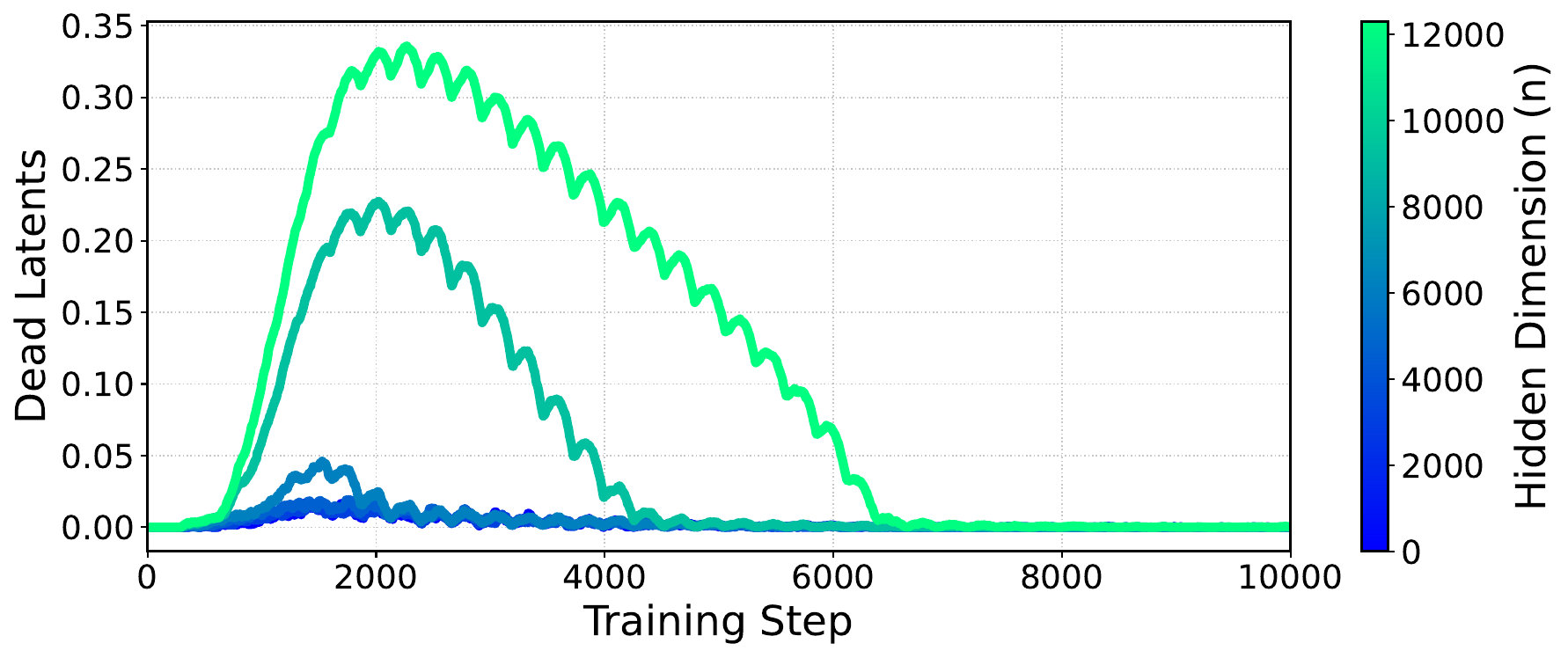}
    \caption{The proportion of dead latents, defined as features that haven't fired in the last epoch of training, for our $k=16$ SAEs on the \astroPH{} abstract embeddings. All dead latents were gone by the end of training. We found that dead latents only occurred in $k=16$ autoencoders.}
    \label{fig:dead_latents_proportion_plot}
\end{figure}

We augment the primary loss with an auxiliary component (AuxK), inspired by the ``ghost grads'' approach of \citet{jermyn2024ghost}. This auxiliary term considers the top-$k_{aux}$ inactive latents (typically $k_{aux}=2k$), where inactivity is determined by a lack of activation over a full training epoch. The total loss is formulated as $\mathcal{L} + \alpha \mathcal{L}_{aux}$, with $\alpha$ usually set to 1/32. This mechanism reduces the number of dead latents with minimal computational overhead \citep{gao2024scaling}. We found that dead latents only occurred during training the $k=16$ models, and all dead latents had disappeared by the end of training. We show how dead latents evolved over training the $k=16$ SAEs for the \astroPH{} abstracts in Figure \ref{fig:dead_latents_proportion_plot}.

For optimisation, we employ Adam \citep{kingma2014adam} with $\beta_1=0.9$ and $\beta_2=0.999$, maintaining a constant learning rate. We use gradient clipping. Our training uses batches of 1024 abstracts, with performance metrics showing robustness to batch size variations under appropriate hyperparameter settings.

The primary MSE loss uses a global normalisation factor computed at training initiation, while the AuxK loss employs per-batch normalisation to adapt to evolving error distributions. Following \citet{bricken2023towards}, we apply a gradient projection technique to mitigate interactions between the Adam optimiser and decoder normalisation.

\subsection{SAE training metrics}

Table \ref{tab:sae-reconstruction-error} shows the final training metrics for all combinations of SAEs trained. We note clear trends in normalised MSE, log feature density and activation mean as we vary the number of active latents $k$ and the overall number of latents $n$. 

\begin{table}[htbp]
\centering
\caption{Metrics for our top-$k$ sparse autoencoders with varying $k$ and hidden dimensions, across both astronomy and computer science papers. MSE is normalised mean squared error, Log FD is the mean log density of feature activations, and activation mean is the mean activation value across non-zero features. Note that MSE is normalised.}
\label{tab:sae-reconstruction-error}
\begin{tabular}{@{}ccc*{3}{c@{\hspace{0.5em}}c@{\hspace{0.5em}}c}@{}}
\toprule
& & & \multicolumn{3}{c}{\texttt{astro.ph}} & \multicolumn{3}{c}{\texttt{cs.LG}} \\
\cmidrule(lr){4-6} \cmidrule(l){7-9}
$k$ & $n$ & & MSE & Log FD & Act Mean & MSE & Log FD & Act Mean \\
\midrule
\multirow{5}{*}{16} & 3072 & & 0.2264 & -2.7204 & 0.1264 & 0.2284 & -2.7314 & 0.1332 \\
 & 4608 & & 0.2246 & -4.7994 & 0.1350 & 0.2197 & -3.0221 & 0.1338 \\
 & 6144 & & 0.2128 & -3.1962 & 0.1266 & 0.2089 & -3.2299 & 0.1342 \\
 & 9216 & & 0.1984 & -3.4206 & 0.1264 & 0.1962 & -3.4833 & 0.1343 \\
 & 12288 & & 0.1957 & -6.2719 & 0.1274 & 0.1897 & -3.6448 & 0.1347 \\
\midrule
\multirow{5}{*}{32} & 3072 & & 0.1816 & -2.3389 & 0.0847 & 0.1831 & -2.3008 & 0.0885 \\
 & 4608 & & 0.1691 & -3.6091 & 0.0882 & 0.1697 & -2.5152 & 0.0876 \\
 & 6144 & & 0.1604 & -2.7761 & 0.0841 & 0.1641 & -2.6687 & 0.0873 \\
 & 9216 & & 0.1554 & -3.0227 & 0.0842 & 0.1540 & -2.9031 & 0.0875 \\
 & 12288 & & 0.1520 & -4.9505 & 0.0843 & 0.1457 & -3.0577 & 0.0877 \\
\midrule
\multirow{5}{*}{64} & 3072 & & 0.1420 & -1.9538 & 0.0566 & 0.1485 & -1.8875 & 0.0584 \\
 & 4608 & & 0.1331 & -2.7782 & 0.0622 & 0.1370 & -2.0637 & 0.0570 \\
 & 6144 & & 0.1262 & -2.2828 & 0.0545 & 0.1310 & -2.1852 & 0.0558 \\
 & 9216 & & 0.1182 & -2.4682 & 0.0539 & 0.1240 & -2.3536 & 0.0545 \\
 & 12288 & & 0.1152 & -3.4787 & 0.0583 & 0.1162 & -2.4847 & 0.0548 \\
\midrule
\multirow{5}{*}{128} & 3072 & & 0.1111 & -1.8876 & 0.0483 & 0.1206 & -1.5311 & 0.0399 \\
 & 4608 & & 0.1033 & -2.1392 & 0.0457 & 0.1137 & -1.6948 & 0.0376 \\
 & 6144 & & 0.1048 & -2.2501 & 0.0438 & 0.1076 & -1.8079 & 0.0366 \\
 & 9216 & & 0.0975 & -2.5352 & 0.0409 & 0.0999 & -1.9701 & 0.0348 \\
 & 12288 & & 0.0936 & -2.7025 & 0.0399 & 0.0942 & -2.0858 & 0.0342 \\
\bottomrule
\end{tabular}
\end{table}

\subsection{Scaling laws}

For the left panel of Figure \ref{fig:combined_plots}, which shows the scaling of normalised MSE with the number of total latents $n$, we observe the following power-law relationships:
\begin{align*}
k = 16 &: L(n) = 0.61n^{-0.12} \text{ (astro.ph)}; \ L(n) = 0.67n^{-0.13} \text{ (cs.LG)} \\
k = 32 &: L(n) = 0.49n^{-0.13} \text{ (astro.ph)}; \ L(n) = 0.56n^{-0.14} \text{ (cs.LG)} \\
k = 64 &: L(n) = 0.46n^{-0.15} \text{ (astro.ph)}; \ L(n) = 0.60n^{-0.17} \text{ (cs.LG)} \\
k = 128 &: L(n) = 0.31n^{-0.13} \text{ (astro.ph)}; \ L(n) = 0.51n^{-0.18} \text{ (cs.LG)}
\end{align*}
For the right panel of Figure \ref{fig:combined_plots}, which shows the scaling of normalised MSE with the amount of compute $C$ (in FLOPs), we observe the following power-law relationships:
\begin{align*}
k = 16 &: L(C) = 3.84C^{-0.11} \\
k = 32 &: L(C) = 5.25C^{-0.13} \\
k = 64 &: L(C) = 8.03C^{-0.16} \\
k = 128 &: L(C) = 2.80C^{-0.13}
\end{align*}
These equations demonstrate the consistent power-law scaling behaviour of sparse autoencoders across different values of $k$, $n$, and compute $C$.

\subsection{Feature density and similarity}

\begin{figure}
\centering
\includegraphics[width=0.7\linewidth]{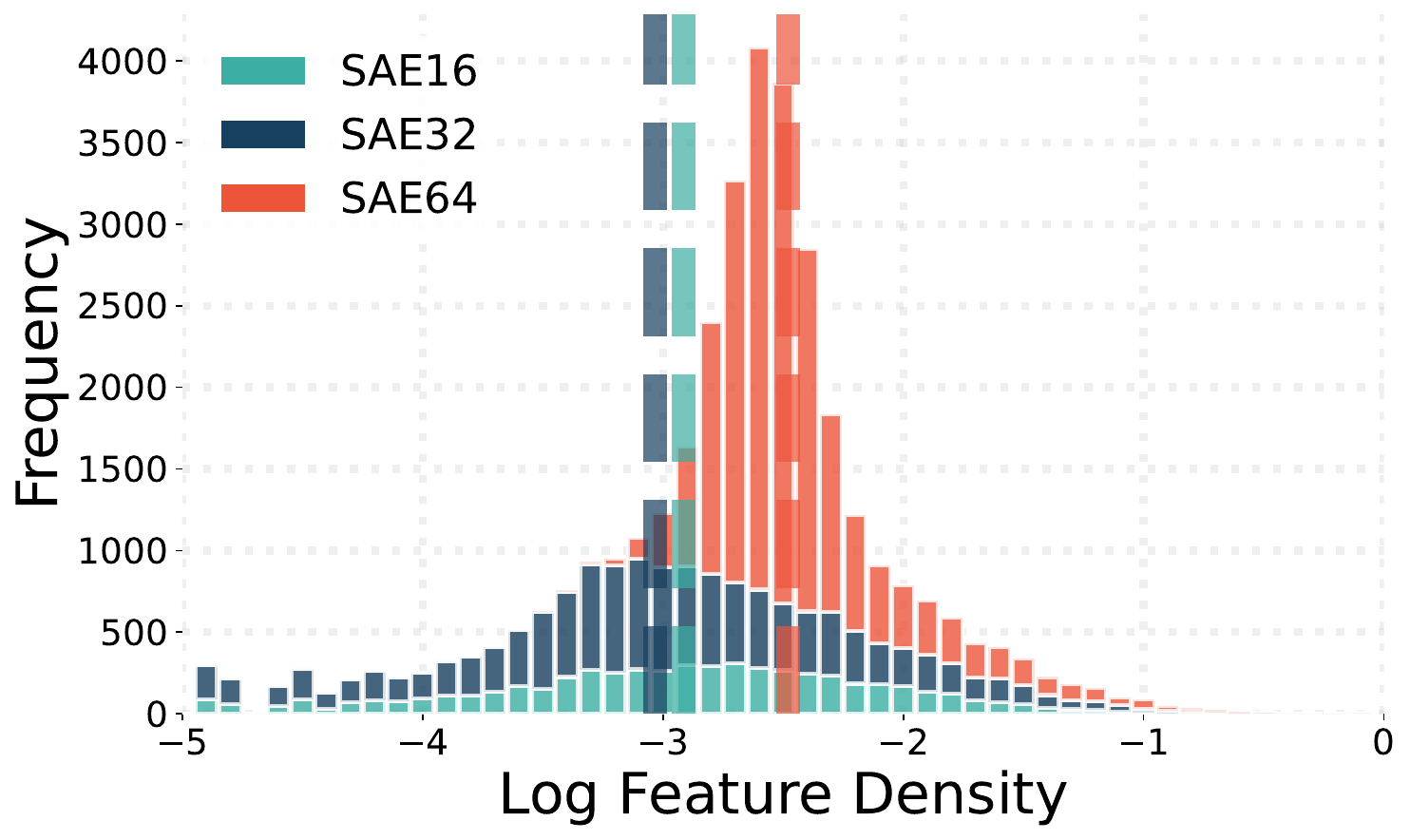}
\caption{Log feature density for features in our three SAEs as a stacked histogram, showing the distribution of how often features fire across all paper abstacts (\csLG{} and \astroPH). The larger SAE has a higher mean feature density than the smaller SAEs.}
\label{fig:log_feature_density}
\end{figure}

We find an intuitive relationship between $k$ and $n$ and the log feature density (essentially, how often a given feature fires). As $k$ increases, we get a sharper peak of log feature density, shifted to the right, suggesting features fire in a tighter range as we increase the instantaneous L0 of the SAE's encoder (Figure \ref{fig:log_feature_density}).

To compare features across different SAEs trained on the same input data, we analyse the cosine similarity between the decoder weight vectors corresponding to each feature. Decoder weights, represented by columns in the decoder matrix, directly encode each feature's contribution to input reconstruction. Encoder weights, on the other hand, are optimised to extract feature coefficients while minimising interference between non-orthogonal features. This separation is important in the context of superposition, where we have more features than input dimensions, precluding perfect orthogonality.

\begin{figure}
    \centering
    \begin{subfigure}[b]{0.7\textwidth}
        \centering
        \includegraphics[width=\textwidth]{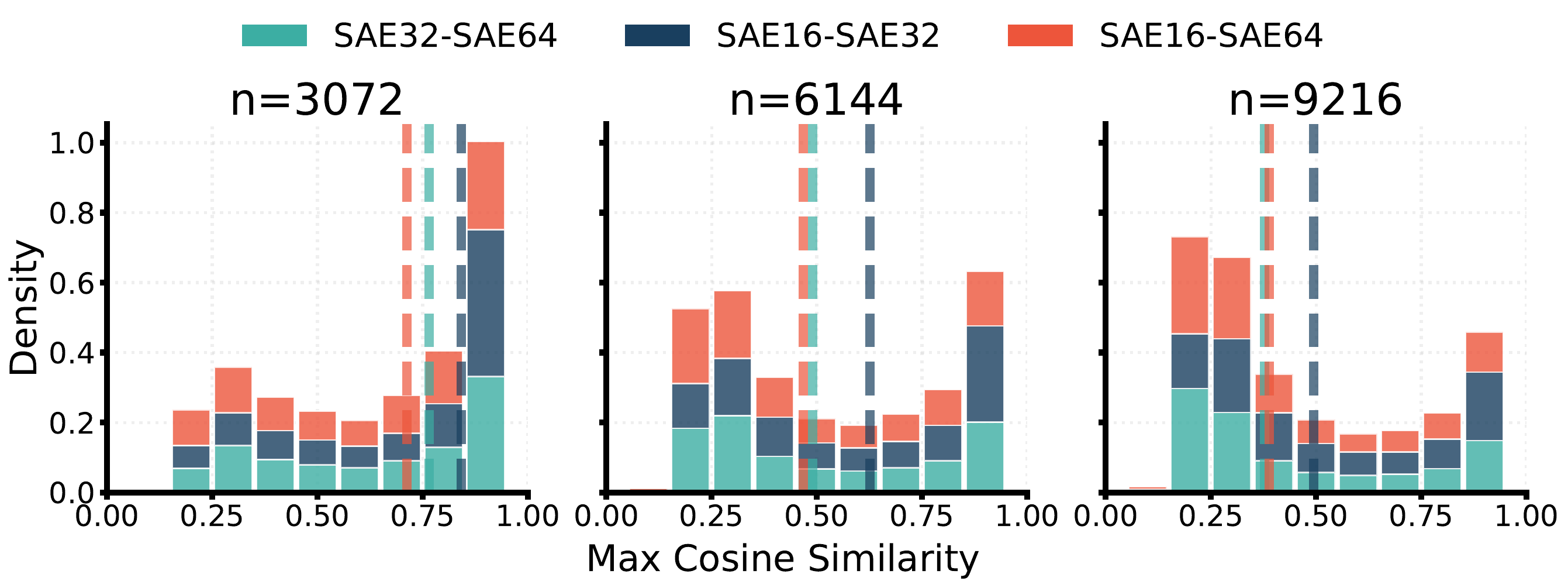}
        \caption{$k$ fixed, varying $n$. As $n$ increases, the features between across SAEs with varying $k$ become more disparate.}
        \label{subfig:combined_cosine_sim_histograms_n}
    \end{subfigure}
    \vspace{1em}
    \begin{subfigure}[b]{0.7\textwidth}
        \centering
        \includegraphics[width=\textwidth]{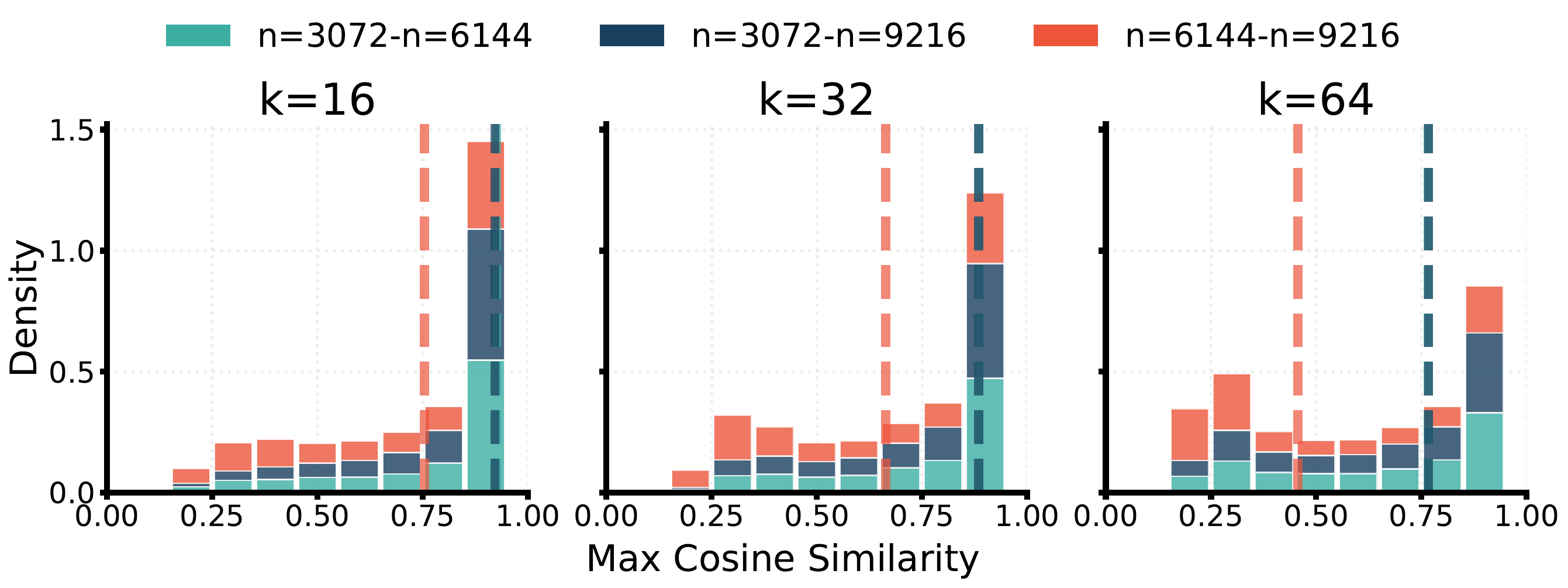}
        \caption{$n$ fixed, varying $k$. Higher values of $k$ lead to less similarity regardless of $n$.}
        \label{subfig:combined_cosine_sim_histograms_k}
    \end{subfigure}
    \caption{Nearest-neighbour cosine similarity distributions for SAE features. To find features in an SAE with a lower $k$ that are most similar to those in an SAE with a larger $k$, we compute the cosine similarity between each feature in the larger model and each feature in the smaller model. We do this for several values of $n$, and combine the distributions for \texttt{astro.ph} and \texttt{cs.LG}.}
    \label{fig:combined_cosine_sim_histograms}
\end{figure}

\section{\texttt{SAErch.ai}}

To demonstrate the practical applications of our sparse autoencoder (SAE) approach to semantic search and feature interpretation, we developed \href{https://huggingface.co/spaces/charlieoneill/saerch.ai}{SAErch.ai}, a web application that allows users to interact with the SAE models trained on arXiv paper embeddings. 

\subsection{Overview}

SAErch.ai is built using the Gradio framework and consists of three main tabs: Home, SAErch, and Feature Visualisation. The application allows users to switch between the Computer Science (\csLG{}) and Astrophysics (\astroPH{}) datasets.

The SAErch tab implements the core functionality of our semantic search system, allowing users to:
\begin{itemize}
\item Input a search query
\item View the top 10 search results based on embedding similarity
\item Interact with the SAE features activated by their query
\end{itemize}

\begin{figure}
    \centering
    \includegraphics[width=1.0\linewidth]{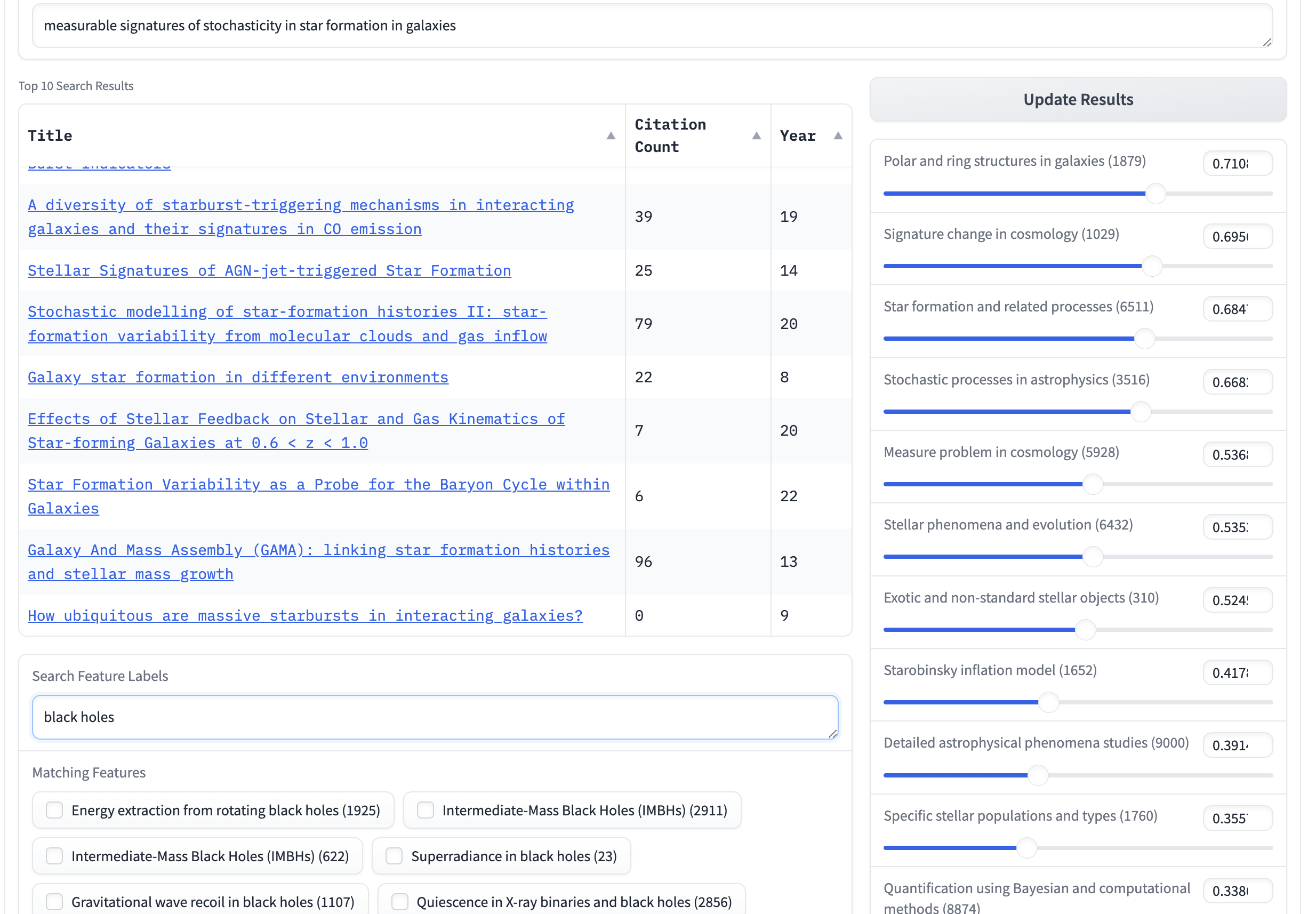}
    \caption{The SAErch tab of our web application, demonstrating a semantic search for ``measurable signatures of stochasticity in star formation in galaxies'' in the astrophysics domain. The interface displays the top 10 search results ranked by relevance, including title, citation count, and publication year. On the right, sliders represent the top activated SAE features for the query, allowing users to fine-tune the search by adjusting feature weights. On the bottom we have our feature addition interface. Users can search for specific semantic features (e.g., ``black holes'') and add them to their query. They can then adjust the strength of these features.}
    \label{fig:saerch_tab_astro}
\end{figure}

For each query, the system displays sliders corresponding to the top-k SAE features activated by the input. Users can adjust these sliders to modify the query embedding, effectively steering the search results towards or away from specific semantic concepts; see Figure \ref{fig:saerch_tab_astro}. This directly demonstrates the fine-grained control over query semantics discussed in Section \ref{sec:saerch} of our paper. Users can also search for and add specific features not initially activated by their query (Figure \ref{fig:add-features-astro}).

\subsection{Feature Visualisation Tab}

The Feature Visualisation tab is divided into two sub-tabs: Individual Features and Feature Families. This section of the application directly relates to our analysis of SAE features and feature families discussed in Sections \ref{sec:training_and_labelling} and \ref{sec:feature_families}.

\subsubsection{Individual Features}

For any selected feature, this tab displays:
\begin{itemize}
\item Top 5 activating abstracts, demonstrating the semantic content captured by the feature
\item Top and bottom 5 correlated features, illustrating the relationships between different SAE features
\item Top 5 co-occurring features, showing which features tend to activate together
\item A histogram of activation values, providing insight into the feature's behavior across the corpus
\item The most similar features in SAE16 and SAE32
\end{itemize}

\begin{figure}
    \centering
    \includegraphics[width=1.0\linewidth]{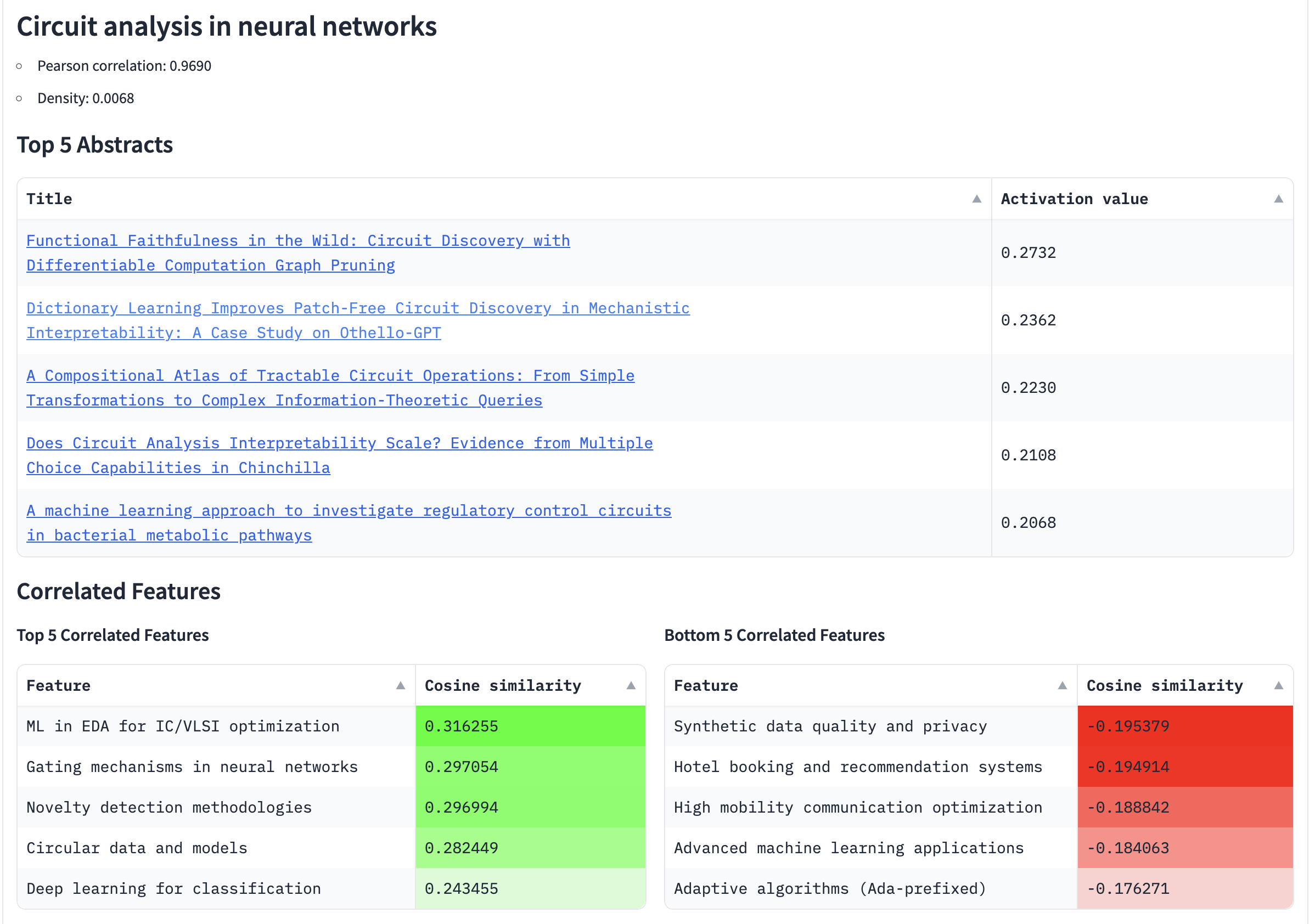}
    \caption{Individual feature visualisation for the ``Circuit analysis in neural networks'' feature in the computer science domain. The interface displays key interpretability metrics, top activating abstracts, correlated and co-occurring features, and an activation distribution histogram. Further information (not shown in the image) includes co-occurring features and activation distribution.}
    \label{fig:individual_features_cs}
\end{figure}




\subsubsection{Feature Families}

\begin{figure}
    \centering
    \includegraphics[width=1.0\linewidth]{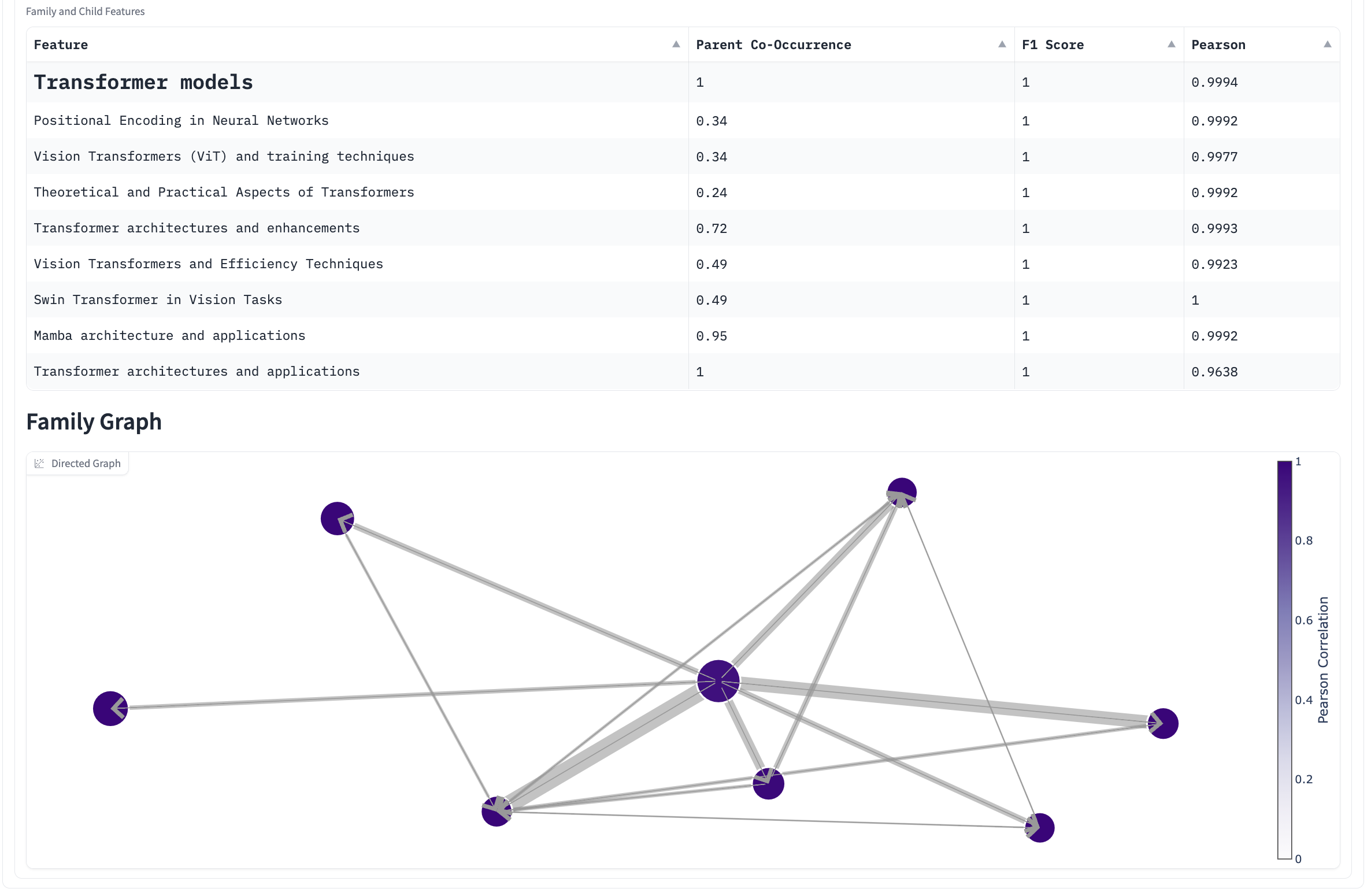}
    \caption{Directed graph visualization of a transformer models feature family. Nodes represent individual features, with size indicating feature density and color intensity showing Pearson correlation. Edges depict relationships between features, with arrow direction pointing from more general to more specific concepts. Users can hover over nodes to view detailed feature information.}
    \label{fig:feat_fam}
\end{figure}

The Feature Families tab in our web application offers an in-depth exploration of related features discovered by our sparse autoencoder. We show an example feature family in Figure \ref{fig:feat_fam}.

The table displays the parent feature (superfeature) and its child features, along with key metrics, such as the name of the parent and child features, the frequency of co-occurrence between the child feature and the parent feature, ranging from 0 to 1, and the F1 Score and Pearson correlation.

The interactive directed graph provides a visual representation of the feature family structure. Each node represents a feature. The size of the node corresponds to the feature's density (frequency of activation), while the color intensity indicates the Pearson correlation (interpretability). Arrows between nodes show relationships between features, with the direction typically pointing from more general to more specific concepts. Users can hover over nodes to view detailed information about each feature, including its name and log density.

\section{Automated interpretability details}
\label{app:automated_interpretability}

\subsection{Examples of features}

\begin{table}[ht]
    \centering
    \scriptsize 
    \begin{tabularx}{\textwidth}{lX X X}
        \toprule
        \textbf{Feature} &  &  &  \\ 
        \midrule
        \multicolumn{4}{l}{\textbf{Astronomy}} \\ \midrule
        Cosmic Microwave Background & \cellcolor{high_green}CMB map-making and power spectrum estimation (0.1708) & \cellcolor{medium_green}How to calculate the CMB spectrum (0.1598) & \cellcolor{medium_green}CMB data analysis and sparsity (0.1581) \\ 
        \midrule
        Periodicity in astronomical data & \cellcolor{low_green}Generalized Lomb-Scargle analysis of decay rate measurements from the Physikalisch-Technische Bundesanstalt (0.1027) & \cellcolor{low_green}Multicomponent power-density spectra of Kepler AGNs, an instrumental artefact or a physical origin? (0.0806) & \cellcolor{low_green}RXTE observation of the X-ray burster 1E 1724-3045. I. Timing study of the persistent X-ray emission with the PCA (0.0758) \\ 
        \midrule
        X-ray reflection spectra & \cellcolor{high_green}X-ray reflection spectra from ionized slabs (0.3859) & \cellcolor{high_green}The role of the reflection fraction in constraining black hole spin (0.3803) & \cellcolor{medium_green}Relativistic reflection: Review and recent developments in modeling (0.3698) \\ 
        \midrule
        Critique or refutation of theories & \cellcolor{high_green} What if string theory has no de Sitter vacua? (0.2917) & \cellcolor{medium_green} No evidence of mass segregation in massive young clusters (0.2051) & \cellcolor{medium_green} Ruling Out Initially Clustered Primordial Black Holes as Dark Matter (0.2029) \\ 
        \midrule
        \multicolumn{4}{l}{\textbf{Computer Science}} \\ \midrule
        Sparsity in Neural Networks & \cellcolor{high_blue}Two Sparsities Are Better Than One: Unlocking the Performance Benefits of Sparse-Sparse Networks (0.3807) & \cellcolor{high_blue}Truly Sparse Neural Networks at Scale (0.3714) & \cellcolor{medium_blue}Topological Insights into Sparse Neural Networks (0.3689) \\ 
        \midrule
        Gibbs Sampling and Variants & \cellcolor{medium_blue}Herded Gibbs Sampling (0.2990) & \cellcolor{medium_blue}Characterizing the Generalization Error of Gibbs Algorithm with Symmetrized KL information (0.2858) & \cellcolor{low_blue}A Framework for Neural Network Pruning Using Gibbs Distributions (0.2843) \\ 
        \midrule
        Arithmetic operations in transformers & \cellcolor{medium_blue}Arbitrary-Length Generalization for Addition in a Tiny Transformer (0.1828) & \cellcolor{medium_blue}Carrying over algorithm in transformers (0.1803) & \cellcolor{low_blue}Understanding Addition in Transformers (0.1792) \\ 
        \bottomrule
    \end{tabularx}
    \caption{Activation strengths and titles for abstracts related to Astronomy and Computer Science features.}
    \label{tab:features}
\end{table}

We show some examples of perfectly interpretable features (Pearson correlation $>0.99$) in Table \ref{tab:features}. The strength of the activation of the feature on its top 3 activating abstracts is shown in parentheses next to the abstract title.

\newpage

\subsection{Automated interpretability prompts}

We provide the prompts used for the Interpreter model and the Predictor model in the boxes below. Where \texttt{this} text is used, it represents an input to the model. We found that performance significantly increased when including the instruction to use ``Occam's razor'', whereby the simplest feature at the appropriate level of granularity was selected.

\newpage

\begin{tcolorbox}[colback=green!5!white, colframe=green!75!black, title=\textcolor{white}{\textbf{Interpreter Model Prompt}}, fonttitle=\bfseries, coltitle=white]

You are a meticulous \texttt{<type>} researcher conducting an important investigation into a certain neuron in a language model trained on \texttt{<subject>} papers. Your task is to figure out what sort of behaviour this neuron is responsible for -- namely, on what general concepts, features, themes, methodologies or topics does this neuron fire? Here's how you'll complete the task:

\vspace{3mm}

\textbf{INPUT DESCRIPTION}:  You will be given two inputs: 1) Max Activating Examples and 2) Zero Activating Examples.
\begin{enumerate}
\item You will be given several examples of text that activate the neuron, along with a number being how much it was activated. This means there is some feature, theme, methodology, topic or concept in this text that `excites' this neuron.
\item You will also be given several examples of text that don't activate the neuron. This means the feature, topic or concept is not present in these texts.
\end{enumerate}

\textbf{OUTPUT DESCRIPTION:} Given the inputs provided, complete the following tasks.
\begin{enumerate}
\item Based on the \texttt{MAX ACTIVATING EXAMPLES} provided, write down potential topics, concepts, themes, methodologies and features that they share in common. These will need to be specific - remember, all of the text comes from \texttt{subject}, so these need to be highly specific \texttt{subject} concepts. You may need to look at different levels of granularity (i.e. subsets of a more general topic). List as many as you can think of. Give higher weight to concepts more present/prominent in examples with higher activations.
\item Based on the zero activating examples, rule out any of the topics/concepts/features listed above that are in the zero-activating examples. Systematically go through your list above.
\item Based on the above two steps, perform a thorough analysis of which feature, concept or topic, at what level of granularity, is likely to activate this neuron. Use Occam's razor, as long as it fits the provided evidence. Be highly rational and analytical here.
\item Based on step 4, summarise this concept in 1-8 words, in the form \texttt{FINAL: <explanation>}. Do NOT return anything after these 1-8 words.
\end{enumerate}

Here are the max-activating examples:
\texttt{<max activating examples>}

\vspace{3mm}

Here are the zero-activating examples:
\texttt{<zero activating examples>}

\vspace{3mm}

Work through the steps thoroughly and analytically to interpret our neuron.

\end{tcolorbox}

\newpage

\begin{tcolorbox}[colback=blue!5!white, colframe=blue!75!black, title=\textcolor{white}{\textbf{Predictor Model Prompt}}, fonttitle=\bfseries, coltitle=white]

You are a \texttt{<subject>} expert that is predicting which abstracts will activate a certain neuron in a language model trained on \texttt{<subject>} papers.  Your task is to predict which of the following abstracts will activate the neuron the most. Here's how you'll complete the task:

\vspace{3mm}

\textbf{INPUT DESCRIPTION:} You will be given the description of the type of paper abstracts on which the neuron activates. This description will be short. You will then be given an abstract. Based on the concept of the abstract, you will predict whether the neuron will activate or not.

\vspace{3mm}

\textbf{OUTPUT DESCRIPTION:} Given the inputs provided, complete the following tasks.
\begin{enumerate}
\item Based on the description of the type of paper abstracts on which the neuron activates, reason step by step about whether the neuron will activate on this abstract or not. Be highly rational and analytical here. The abstract may not be clear cut - it may contain topics/concepts close to the neuron description, but not exact. In this case, reason thoroughly and use your best judgement. However, do not speculate on topics that are not present in the abstract.
\item Based on the above step, predict whether the neuron will activate on this abstract or not. If you predict it will activate, give a confidence score from 0 to 1 (i.e. 1 if you're certain it will activate because it contains topics/concepts that match the description exactly, 0 if you're highly uncertain). If you predict it will not activate, give a confidence score from -1 to 0.
\item Provide the final confidence score in the form \texttt{PREDICTION: (your prediction)} e.g. \texttt{PREDICTION: 0.5}. Do NOT return anything after this.
\end{enumerate}

Here is the description/interpretation of the type of paper abstracts on which the neuron activates:
\texttt{<description>}

\vspace{3mm}

Here is the abstract to predict:
\texttt{<abstract>}

\vspace{3mm}

Work through the steps thoroughly and analytically to predict whether the neuron will activate on this abstract.

\end{tcolorbox}

\subsection{Exploring the effectiveness of smaller models}

\begin{figure}
    \centering
    \includegraphics[width=1.0\linewidth]{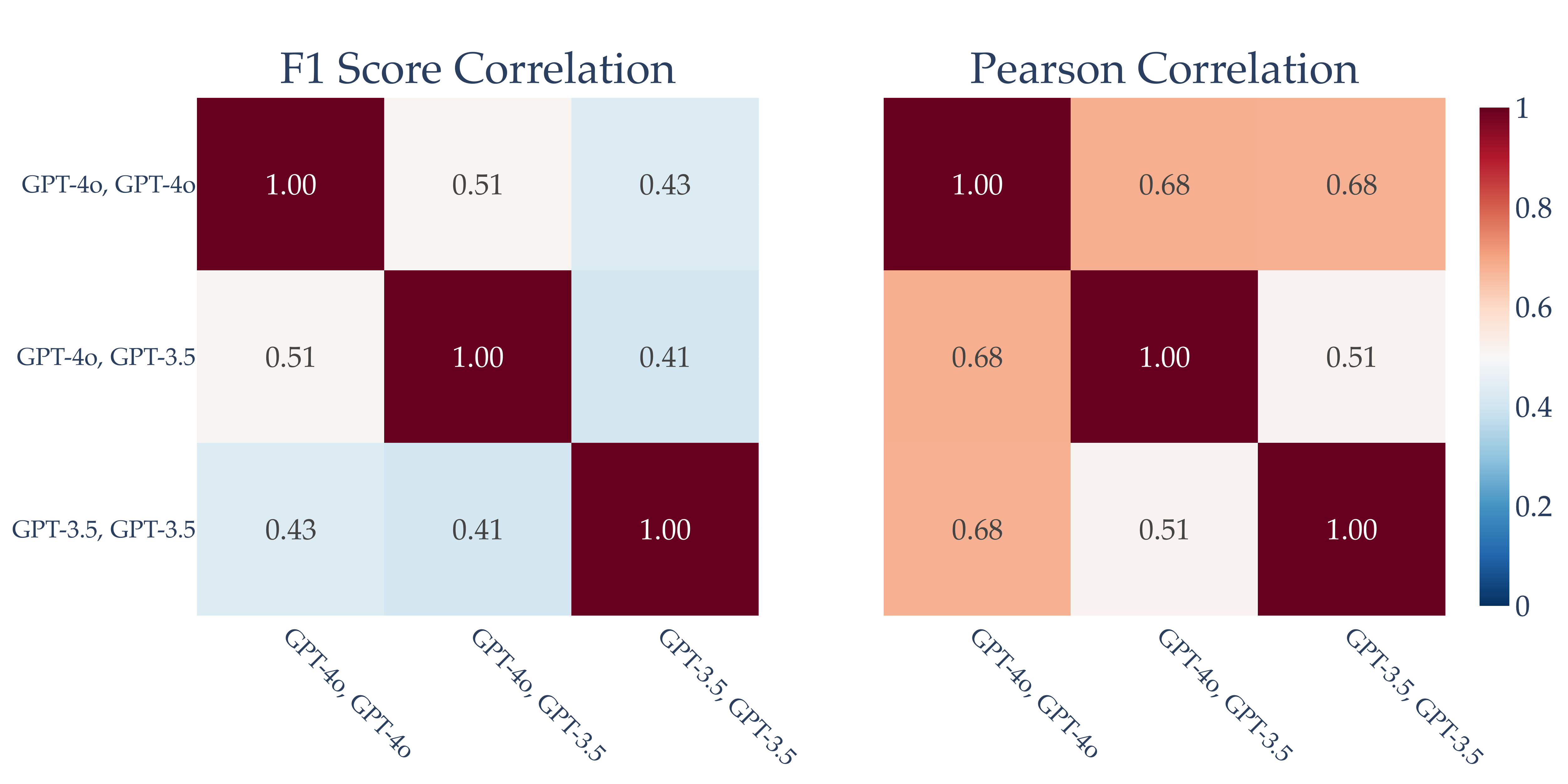}
    \caption{Correlation between F1 scores and Pearson correlation scores of different combinations of \texttt{(labeller, predictor)} models. Interestingly, using GPT-3.5 as the predictor appears to degrade performance similarly regardless of whether the feature was labelled by GPT-4o or GPT-3.5.}
    \label{fig:autointerp_correlation_heatmaps}
\end{figure}

\begin{figure}
    \centering
    \includegraphics[width=1.0\linewidth]{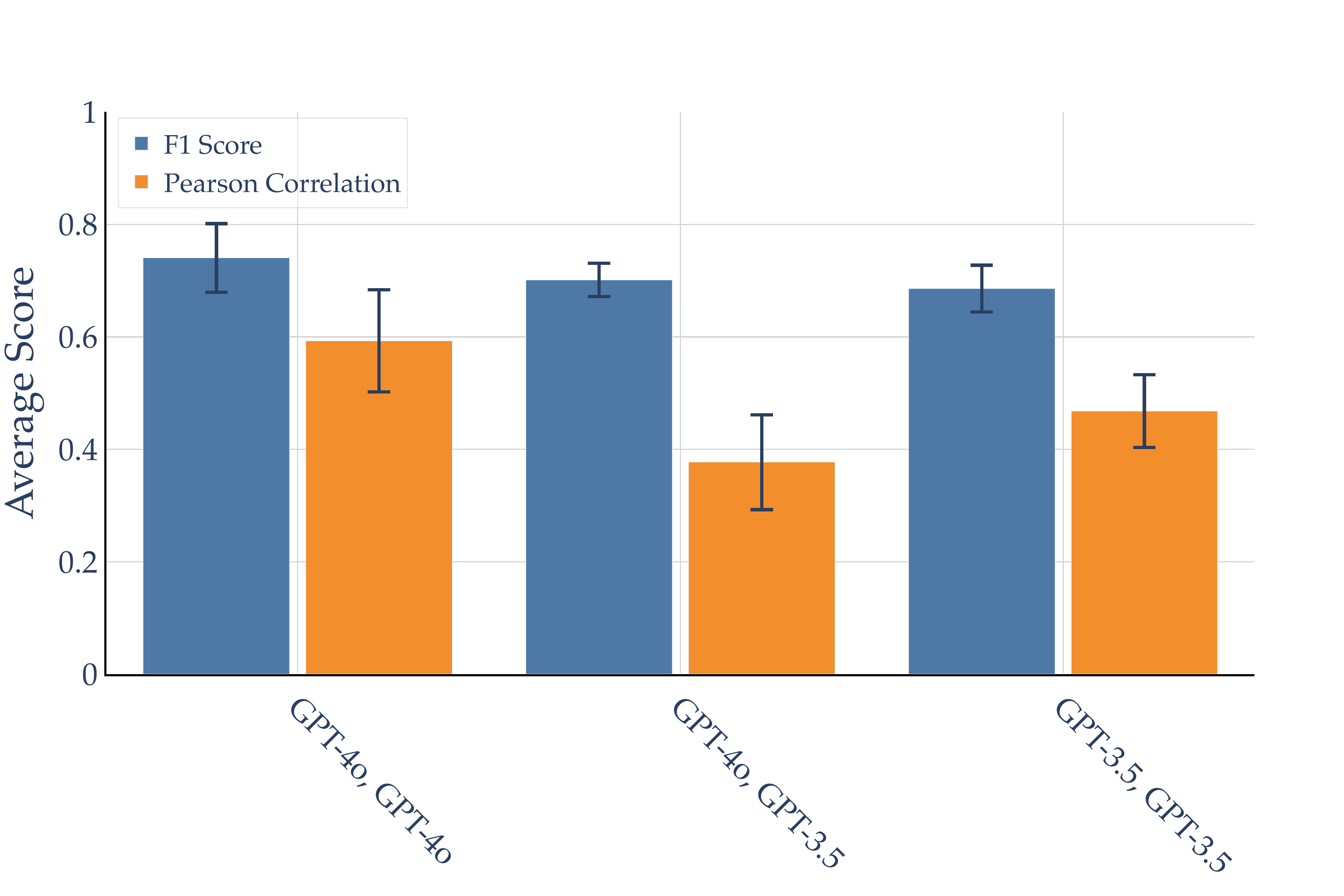}
    \caption{Mean F1 scores and Pearson correlations (according to ground-truth feature activations) across 50 randomly sampled features, for different combinations of \texttt{(Interpreter, Predictor)} models.}
    \label{fig:average_metrics_bar_plot}
\end{figure}

Although we eventually used \texttt{gpt-4o-mini} as the Predictor model, we initially did some ablations to understand how effective \texttt{gpt-4o} and \texttt{gpt-3.5-turbo} would be as different combinations of the Interpreter and Predictor models. We measured this by randomly sampling 50 features from our SAE64 (trained on \astroPH{} abstracts) and measuring the interpretability scores of different model combinations, in terms of both F1 score (does the model's binary classification of a feature firing on an abstract agree with the ground-truth) and the Pearson correlation (described in the main body). Interestingly, we observe that using \texttt{gpt-4o} as the Interpreter and \texttt{gpt-3.5-turbo} as the Predictor leads to similar scores as using \texttt{gpt-3.5-turbo} for both, as shown in Figures \ref{fig:autointerp_correlation_heatmaps} and Figures \ref{fig:average_metrics_bar_plot}. This suggests that the challenging task in the autointerp is not necessarily labelling but rather predicting the activation of a feature on unseen abstracts.

Another observation is that using \texttt{gpt-3.5-turbo} as the Predictor only leads to a moderate degradation of F1 score, it leads to a significant degradation of Pearson correlation. This is likely because we only use 6 abstracts for each feature prediction (3 positive, 3 negative) and thus there are only a few discrete F1 scores possible. Additionally, it appeared that \texttt{gpt-3.5-turbo} was generally less likely to assign higher confidence scores in either direction, with a much lower variance in assigned confidence than when \texttt{gpt-4o} was the Predictor. This affects Pearson correlation but not F1.

\section{Cross-domain features}
\begin{figure}
    \centering
    \begin{subfigure}[b]{0.9\textwidth}
        \centering
        \includegraphics[width=\textwidth]{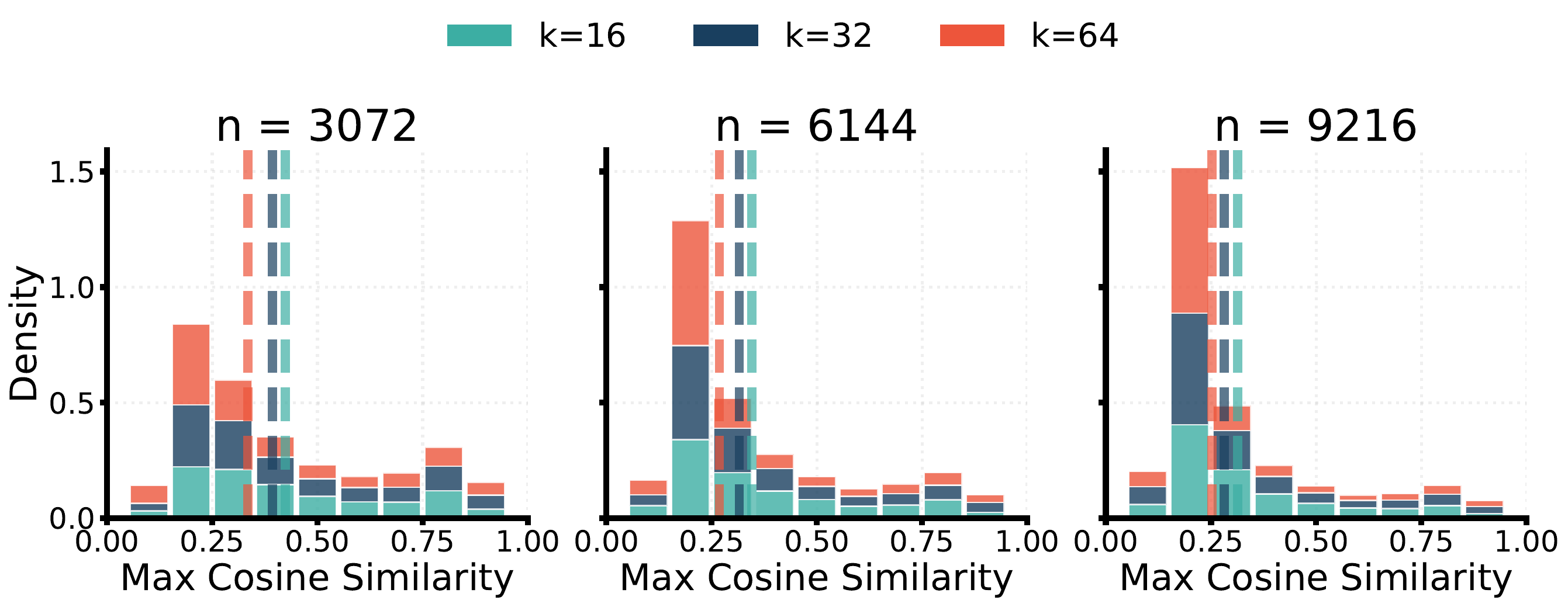}
        \label{fig:cs-astro-hist}
    \end{subfigure}
    \hfill
    \caption{Maximum pair-wise cosine similarity of feature vectors between SAEs trained on different domains.}
    \label{fig:cs-astro}
\end{figure}

The intersection between our \csLG{} ($n = 153,146$) and \texttt{astro.PH} ($n = 271,492$) corpora contains $n = 330$ cross-posted papers. Motivated by these papers, as well as the observation of similar features re-occurring in models of different sizes (see Section \ref{sec:feature_families}), we search for the max cosine similarity feature between \csLG{} and \texttt{astro.PH} SAEs at a fixed $k$ and $n_{dir}$. As expected, we find significant mis-alignment between the vast majority of feature vectors between SAEs trained on different domains, with mis-alignment increasing with $k$ and $n_{dir}$ (see Figure \ref{fig:cs-astro}; this is unsurprising given how $k$ and $n_{dirs}$ correlate with feature granularity). 

However, a small subset of features appear in both sets of SAEs, with relatively high max cosine similarity. For example, Table \ref{fig:cs-astro-table} shows the nearest \csLG{} neighbours for every feature in the \texttt{astro.PH} ``Machine Learning'' feature family (average cosine similarity = $0.59$, average activation similarity = $0.40$). To test whether the features represent the same semantic concepts, we substitute the natural language description of the best-match \csLG{} feature for each listed \texttt{astro.PH} feature and test the interpretability of the substituted descriptions; we find $\Delta_{\text{Pearson}} = -0.07$ and $\Delta_{F1} = -0.06$. The existence of these features suggests that both sets of SAEs learn a semi-universal set of features that span the domain overlap between \texttt{astro.PH} and \csLG{}.

Interestingly, we find a number of near-perfectly aligned pairs (cosine similarity $> 0.95$) of highly interpretable features with little semantic overlap. A number of these features share similar wording but not meaning, such as ``Substructure in dark matter and galaxies" (\astroPH{}) and ``Subgraphs and their representations". Of these 10 feature pairs, the average activation similarity is 0.91. 

\begin{table}[]
        \centering
        \scriptsize
\begin{tabular}{llrrrr}
\hline
Feature Name (\astroPH{}) & Best Match (\csLG{}) & Cosine Sim. & Activation Sim. & $\Delta$ F1 & $\Delta$ Pearson \\
 Deep learning                          & CNNs and Applications &          0.39 &              0.33 &         -0.2  &              -0.17 \\
 Generative Adversarial Networks & Generative Adversarial Networks (GANs)         &          0.61 &              0.26 &          0    &               0    \\
 Transformers                           & Transformer architectures and applications     &          0.5  &              0.33 &          0    &              -0    \\
 Artificial Neural Networks             & Artificial Neural Networks (ANNs)              &          0.64 &              0.02 &          0    &               0    \\
 Artificial Intelligence                & AI applications in diverse domains             &          0.61 &              0.45 &          0    &               0.02 \\
 Automation and Machine Learning        & Automation in computational processes          &          0.9  &              0.77 &         -0.25 &              -0.47 \\
 Gaussian Processes                     & Gaussian Processes in Machine Learning         &          0.59 &              0.54 &          0    &               0.03 \\
 Regression analysis                    & Regression techniques and applications         &          0.81 &              0.53 &          0    &              -0.01 \\
\hline
\end{tabular}

        \caption{Feature matches from the "Machine Learning" family (\texttt{astroPH}); $k = 64, n_{dir} = 9216$. }
        \label{fig:cs-astro-table}
\end{table}

\section{Feature family details}

\subsection{Feature splitting structures} \label{section:family}
\begin{figure}
    \centering
    \includegraphics[width=0.5\linewidth]{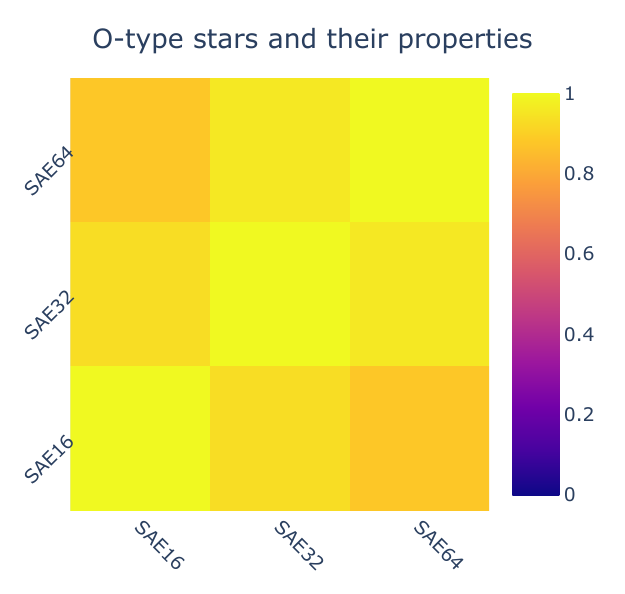}
    \caption{Recurrent features across SAEs trained on \astroPH{}; heatmap colored by activation similarity $D$; all feature vector cosine similarities are $> 0.98$.}
    \label{fig:splitting}
\end{figure}

\begin{figure} 
    \centering
    \begin{subfigure}[t]{0.7\textwidth}
        \centering
        \includegraphics[width=\linewidth]{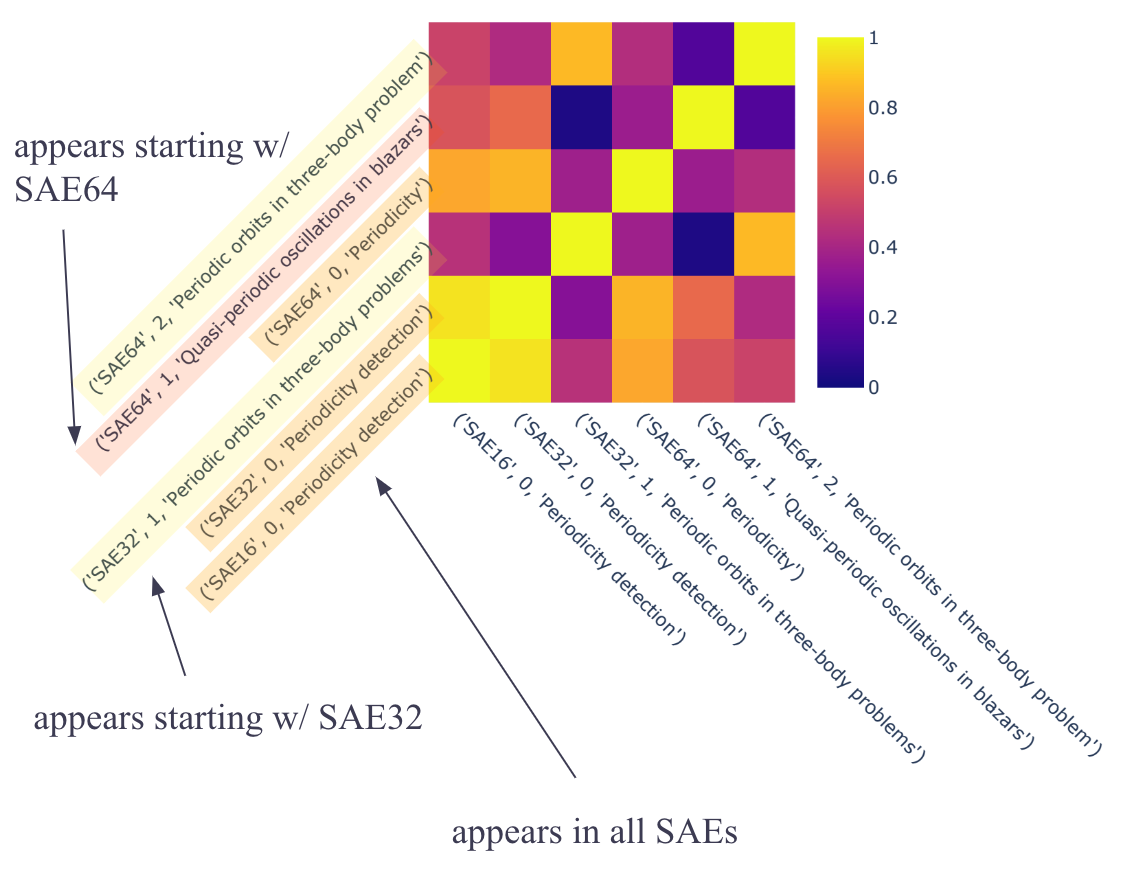} 
        \caption{We find both recurrent features and novel features at every level (i.e. the top-level ``periodicity detection"/``periodicity" feature); heatmap colored by pairwise cosine similarity.} \label{fig:splitting2}
    \end{subfigure}
    \hfill
    \begin{subfigure}[t]{0.8\textwidth}
        \centering
        \includegraphics[width=\linewidth]{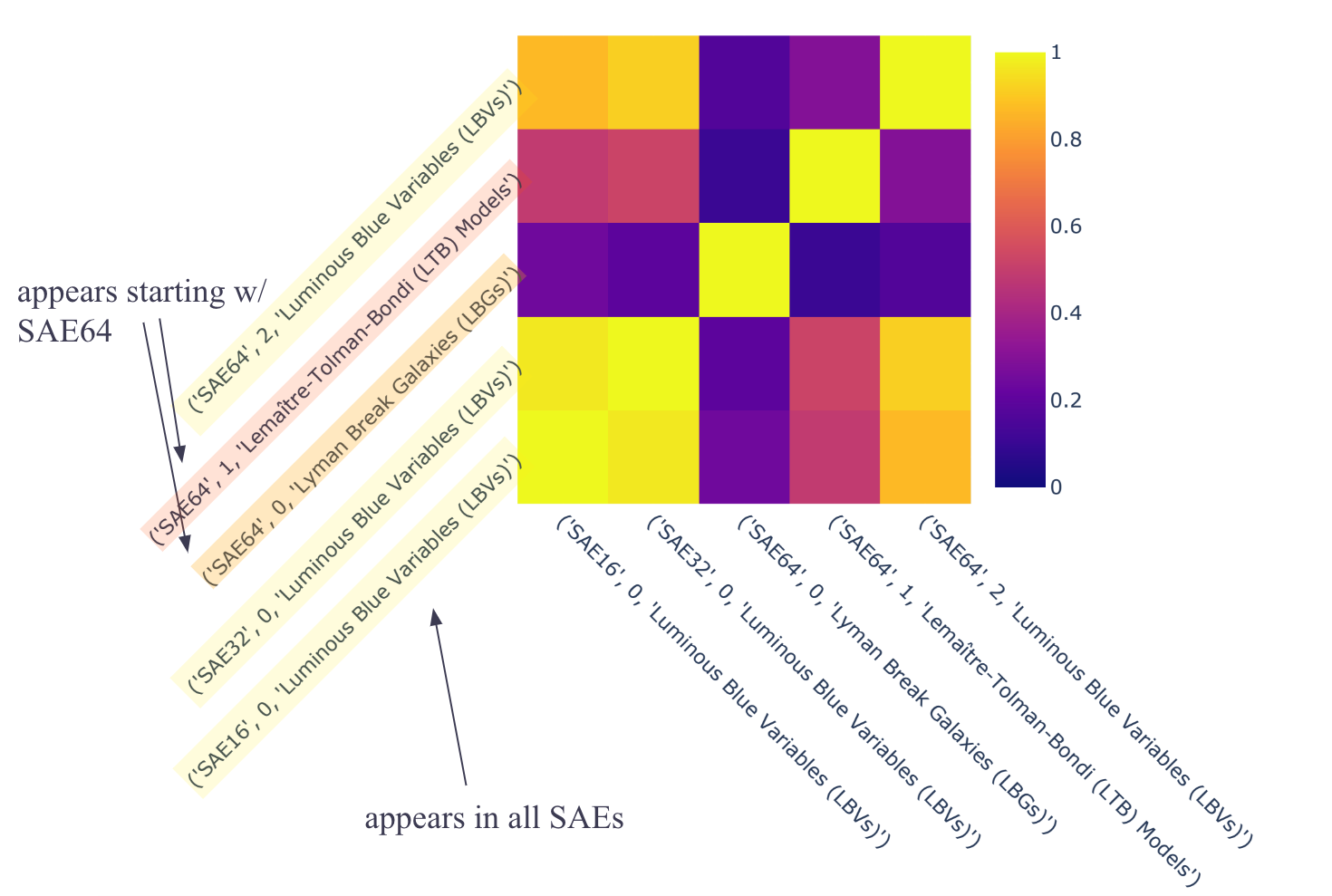} 
        \caption{While ``Luminous Blue Variables" is a recurrent feature in each SAE, SAE64 also exhibits 2 other nearest-neighbour features to ``Luminous Blue Variables" that are not semantically related; heatmap colored by pairwise cosine similarity.} \label{fig:splitting3}
    \end{subfigure}
    \label{fig:splitting4}
\end{figure}

Figure \ref{fig:splitting} shows an example of a recurrent feature across SAE sizes that does not exhibit feature splitting. While the feature has extremely high activation and cosine similarity across every model pair, each model only learns 1 feature in this direction. In Figures \ref{fig:splitting2} and \ref{fig:splitting3} we show two examples of feature splitting across SAE16 -- SAE32 -- SAE64 trained on \astroPH{}. \ref{fig:splitting2} appears to show canonical feature splitting as originally described in \cite{bricken2023towards}, with an increasing number of features splitting the semantic space at each SAE size. 
There exists a top-level ``periodicity''/``periodicity detection'' feature universal to all three SAEs, with relatively high similarity to all other features, as well as novel, more granular features appearing in smaller SAEs, i.e. ``Quasi-periodic oscillations in blazars'', which only appears in SAE64 and is highly dissimilar from other split features.

In contrast, \ref{fig:splitting3} demonstrates nearest-neighbour features across models that do not exhibit semantically meaningful feature splitting. While the top-level ``Luminous Blue Variables (LBVs)" feature occurs at every model size, SAE64 also exhibits two additional features, ``Lemaitre-Tolman-Bondi (LTB) Models" and ``Lyman Break Galaxies (LBGs)", that are highly dissimilar to each other, the LBVs feature, and every other feature in the smaller models. We claim these are novel features, occurring for the first time in SAE64, and that SAE16/SAE32 do not learn features for any related higher-level concepts; instead, this grouping could be a spurious token-level correlation (LBV/LTB/LBG as similar acronyms).

\begin{figure}
    \centering
    \begin{subfigure}[b]{0.53\textwidth}
        \includegraphics[width=\linewidth]{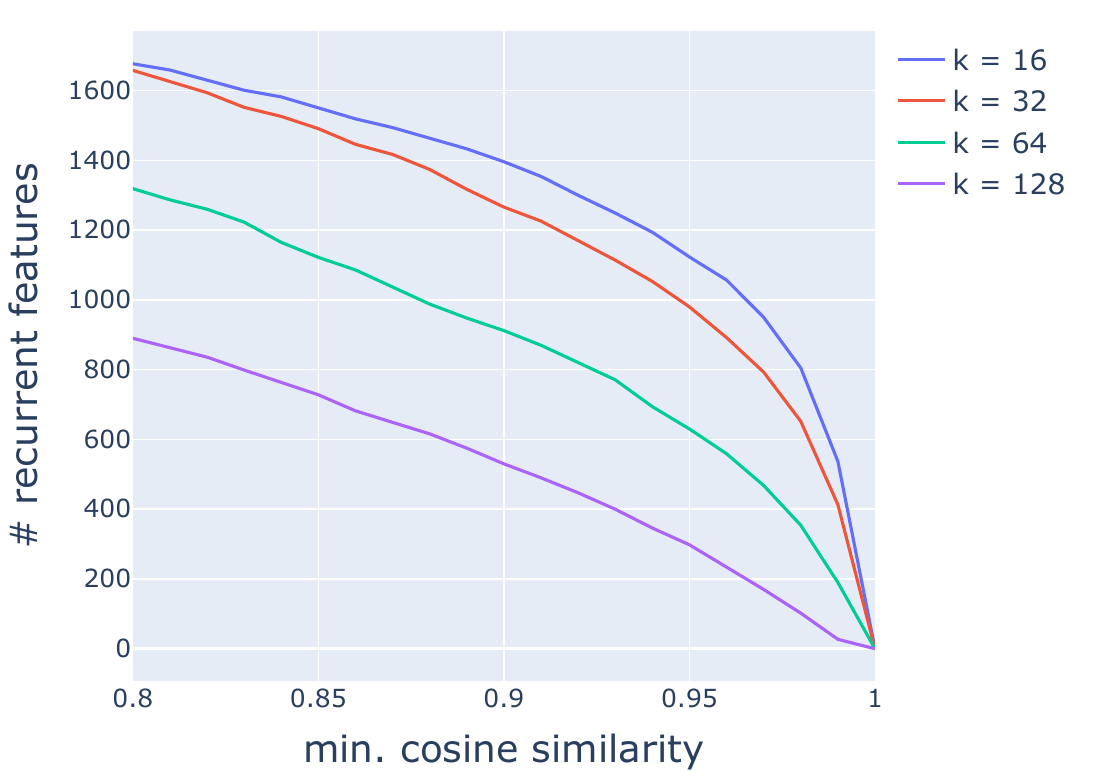}
    \caption{Number of features from the smallest SAE that re-occur in all SAEs, by cosine similarity threshold.}
    \label{fig:survive}
    \end{subfigure}
    \hfill
    \centering
    \begin{subfigure}[b]{0.42\textwidth}
        \includegraphics[width=\linewidth]{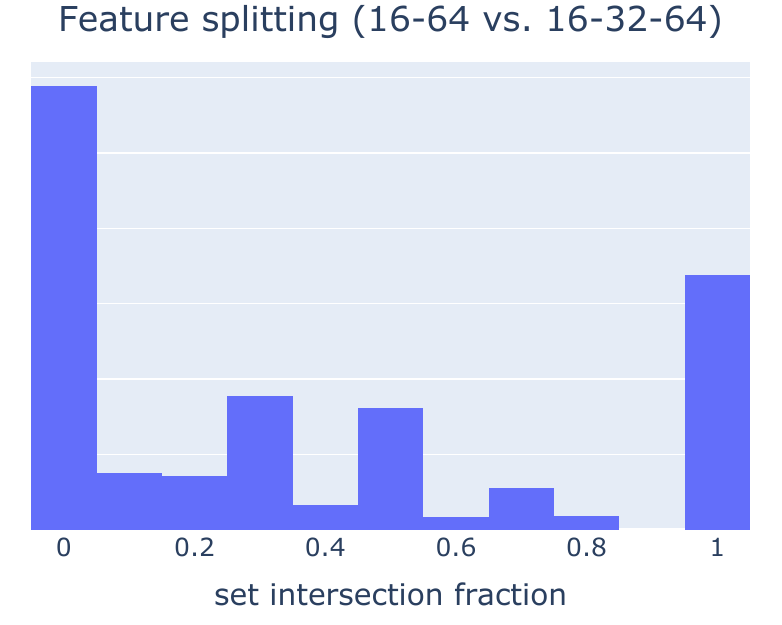}
    \caption{Overlap in the recovered SAE64 features, propagating nearest neighbors from SAE16-SAE64 vs. SAE16-SAE32-SAE64.}
    \label{fig:splitconsistency}
    \end{subfigure}
    
\end{figure}

\paragraph{Feature triplets} In Figure \ref{fig:survive}, we search for features that occur in $n_{dirs} = 3072$ models and have highly aligned features in larger ($n_{dirs} = 6144, 9216$) models; we use this as a rough proxy for the number of re-occurring features. We find that significantly more features re-occur between models for higher $k$, with over 1100 feature triplets at $> 0.95$ cosine similarity for $k = 16$; as $k$ increases, the number of triplets drops sharply.

\paragraph{Self-consistency} In \ref{fig:splitconsistency} we show the set overlap between nearest-neighbour matches between SAE16 and SAE64 found directly, and nearest-neighbour matches between SAE16 and SAE64 found via nearest-neighbour matches to SAE32. If features exhibit perfectly clean splitting geometry, then these two sets of SAE64 features should be consistent. However, we find that the distribution of set overlap is roughly bimodal; other than triplet features with perfect overlap, overlap generally ranges from 0 to 0.6.  The vast majority of intersection = 1 sets are $\leq 3$ features in size. This corroborates findings in \ref{fig:combined_cosine_sim_histograms} which suggests  features across models with different $k$ are not well-aligned.

\subsection{Feature family structure}

\begin{figure}
    \centering
    \begin{subfigure}[b]{0.32\textwidth}
        \centering
        \includegraphics[width=\textwidth]{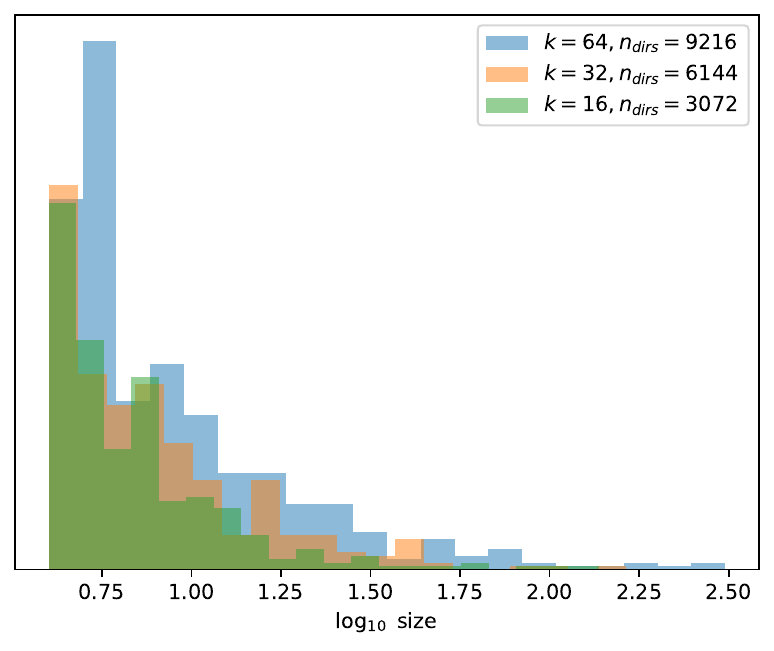}
        \caption{}
        \label{fig:subfig1}
    \end{subfigure}
    \hfill
    \begin{subfigure}[b]{0.32\textwidth}
        \centering
        \includegraphics[width=\textwidth]{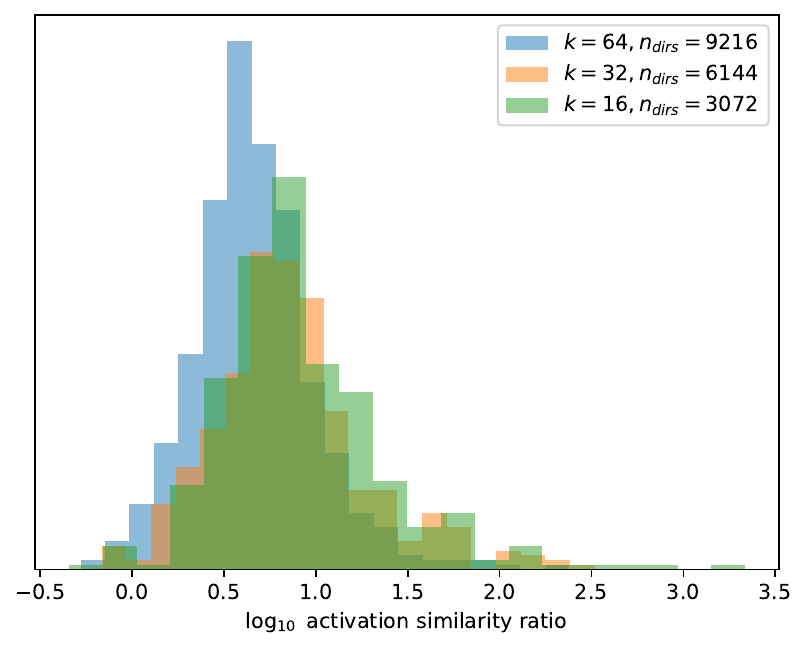}
        \caption{}
        \label{fig:subfig2}
    \end{subfigure}
\hfill
    \begin{subfigure}[b]{0.32\textwidth}
        \centering
        \includegraphics[width=\textwidth]{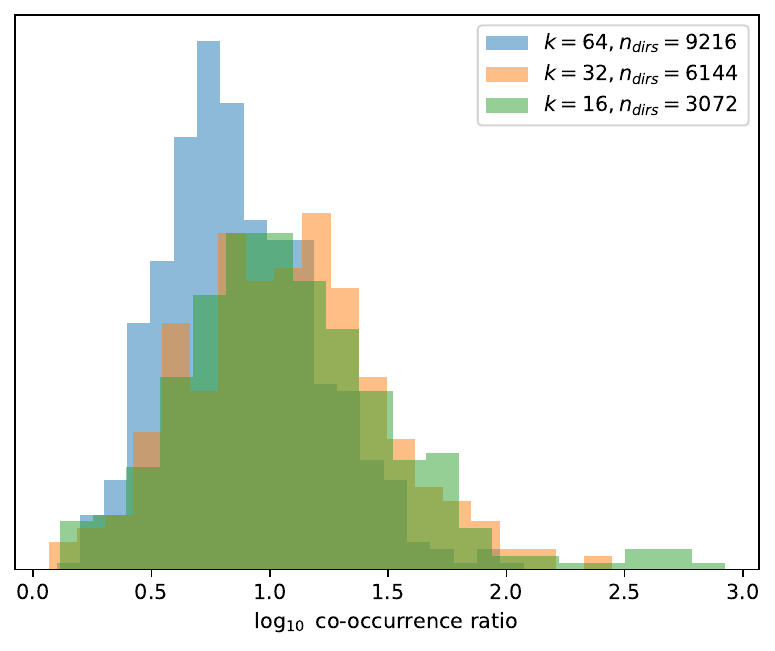}
        \caption{}
        \label{fig:subfig1}
    \end{subfigure}
    \\
    \begin{subfigure}[b]{0.32\textwidth}
        \centering
        \includegraphics[width=\textwidth]{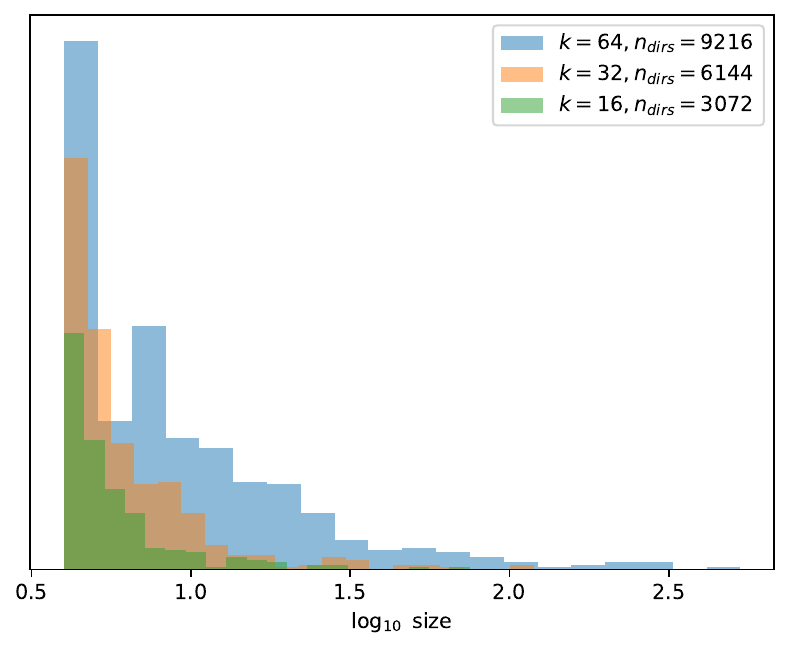}
        \caption{}
        \label{fig:subfig2}
    \end{subfigure}
\hfill
    \begin{subfigure}[b]{0.32\textwidth}
        \centering
        \includegraphics[width=\textwidth]{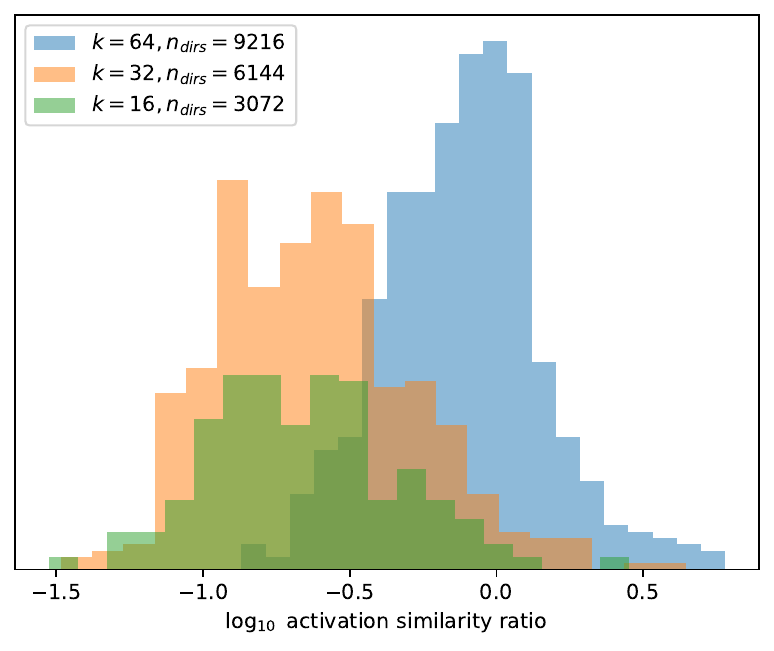}
        \caption{}
        \label{fig:subfig1}
    \end{subfigure}
    \hfill
    \begin{subfigure}[b]{0.32\textwidth}
        \centering
        \includegraphics[width=\textwidth]{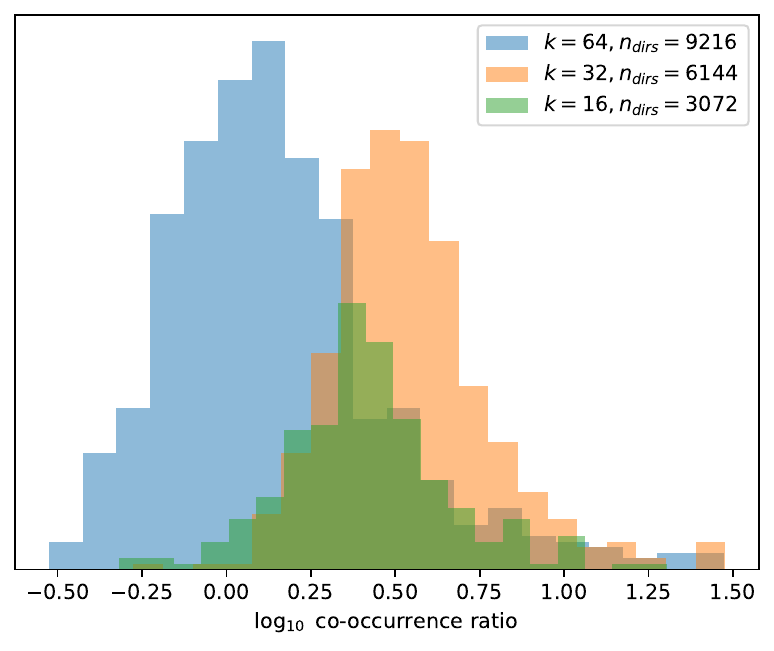}
        \caption{}
        \label{fig:subfig2}
    \end{subfigure}
    
    \caption{Feature families statistics (left: size; middle: activation similarity ratio; right: co-occurrence ratio, $\overline{R(p, \mathcal{C})}$); $k = 64, n_{dir} = 9216$.}
    \label{fig:histograms}
\end{figure}

We compute feature family sizes (including the parent), co-occurrence ratios ($\overline{R(p, \mathcal{C})}$, see section \ref{sec:feature_families}), and activation similarity ratios (computed identically to $\overline{R(p, \mathcal{C})}$, just using activation similarities). Statistics for variants of \csLG{} and \astroPH{} are shown in \ref{fig:histograms}. We find a positive correlation (Spearman = 0.22) between $\overline{R(p, \mathcal{C})}$ and feature family interpretability.

\begin{figure}
    \centering
    \includegraphics[width=0.95\linewidth]{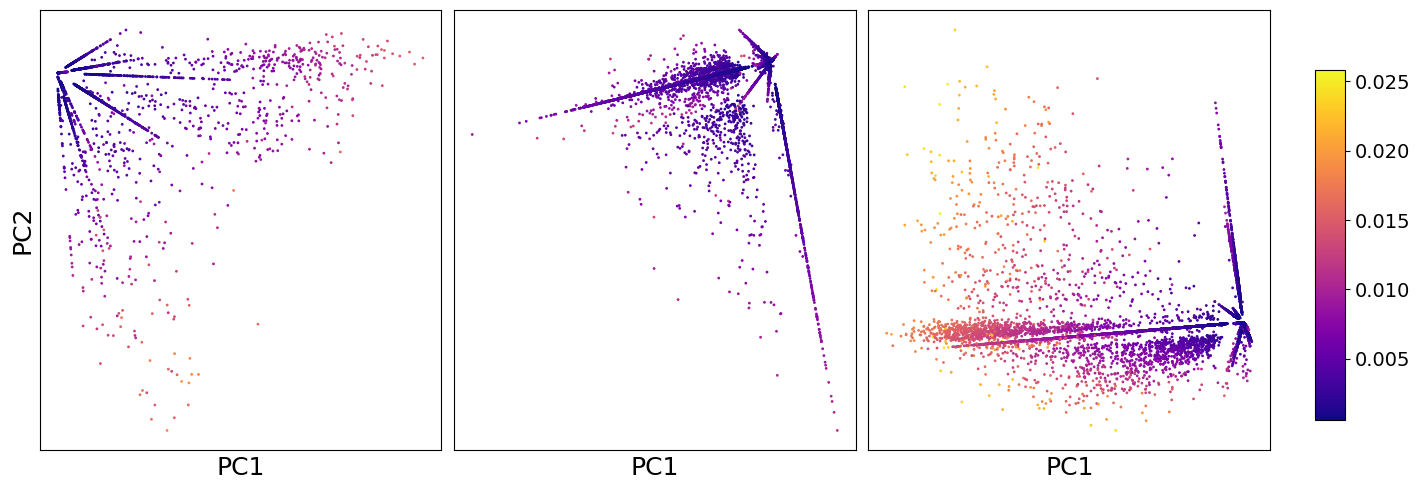}
    \caption{PCA projections of 3 example feature families from SAE64; points are latent representations of activating examples, colored by average activation for in-family features in the top $k$.}
    \label{fig:pca}
\end{figure}

We reproduce the projection method of \cite{engels2024languagemodelfeatureslinear}, running all documents through the SAE and ablating features not in the feature family, to produce Figure \ref{fig:pca}. Visualizing the resulting principal components confirms that the feature families we find do not represent manifolds or irreducible multi-dimensional structures. We can instead think of feature families as linear subspaces in the high-dimensional latent space; in fact, the component vectors can be seen in the lines of points representing documents only activating on one feature in the family.

\begin{figure}
    \centering
    \includegraphics[width=0.7\linewidth]{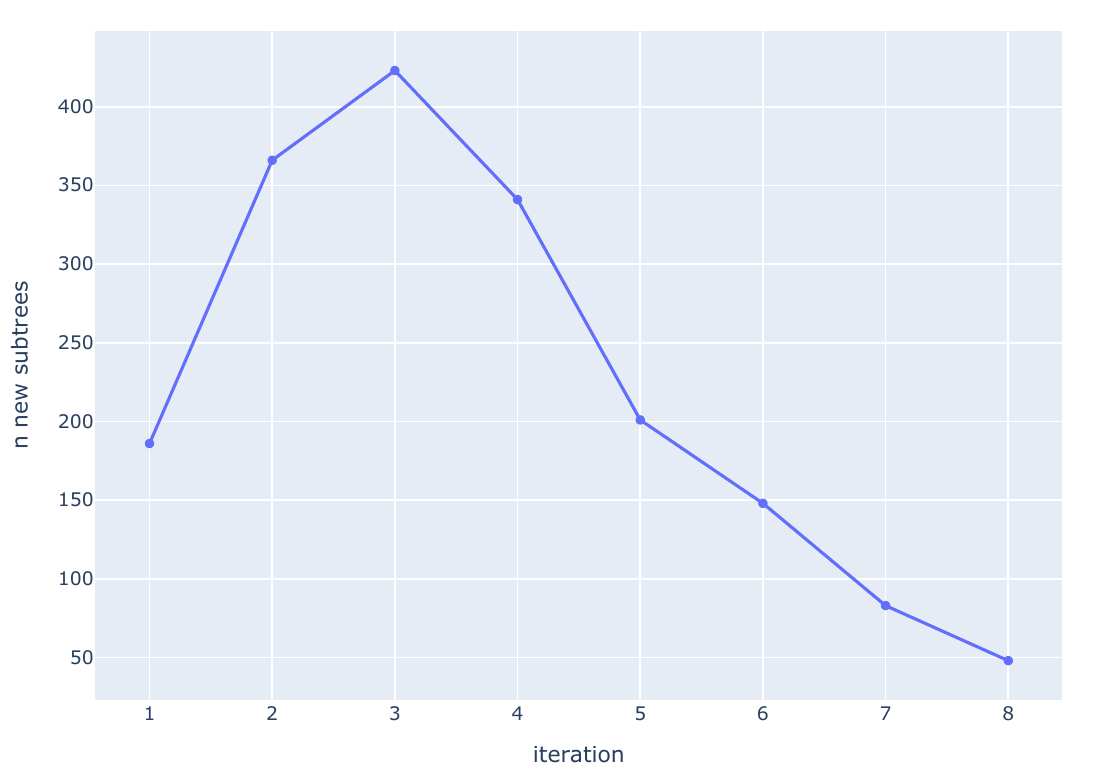}
    \caption{New feature families as a function of iteration; no deduplication is performed.}
    \label{fig:family-iteration}
\end{figure}

In \ref{sec:feature_families} we use $n = 3$ iterations of feature family construction. We select this hyper-parameter based off Figure \ref{fig:family-iteration}. In the first 2-3 iterations, removing parent nodes and re-constructing features preferentially creates additional smaller families, suggesting iterations are necessary to fully explore the graph. But given the sparse co-occurrences ($C_{i, j} > 0.1$) used to build the graph, the number of additional feature families found at each iteration drops off steeply after $n = 3$.

\subsection{Feature family interpretability}
\begin{figure}
    \centering
    \includegraphics[width=1\linewidth]{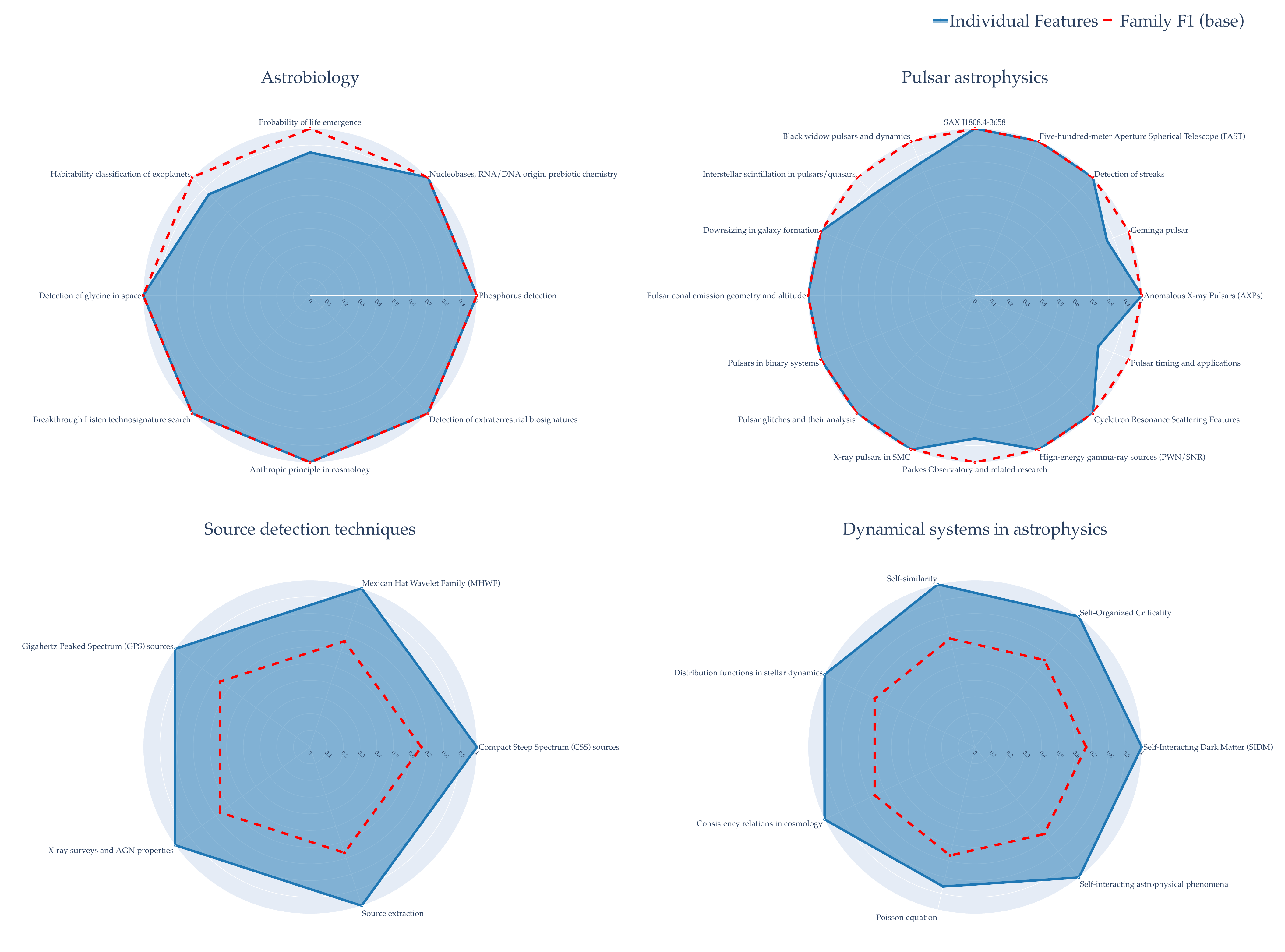}
    \caption{High-quality (top) and low-quality (bottom) feature families, scored through automated interpretability; radar charts show Pearson correlation scores for individual features (vertices) and the overall family (dashed line). While high-quality feature families truly have shared meaning, low-quality families appear to be mostly spurious and are not interpretable through short descriptions.}
    \label{fig:radar_charts}
\end{figure}

We show example feature families and their interpretability scores in Figure \ref{fig:radar_charts}.


\section{Exploring learned decoder weight matrices}
\label{app:decoder_weight_matrices}

\begin{figure}
    \centering
    \begin{minipage}{0.49\linewidth}
        \includegraphics[width=\linewidth]{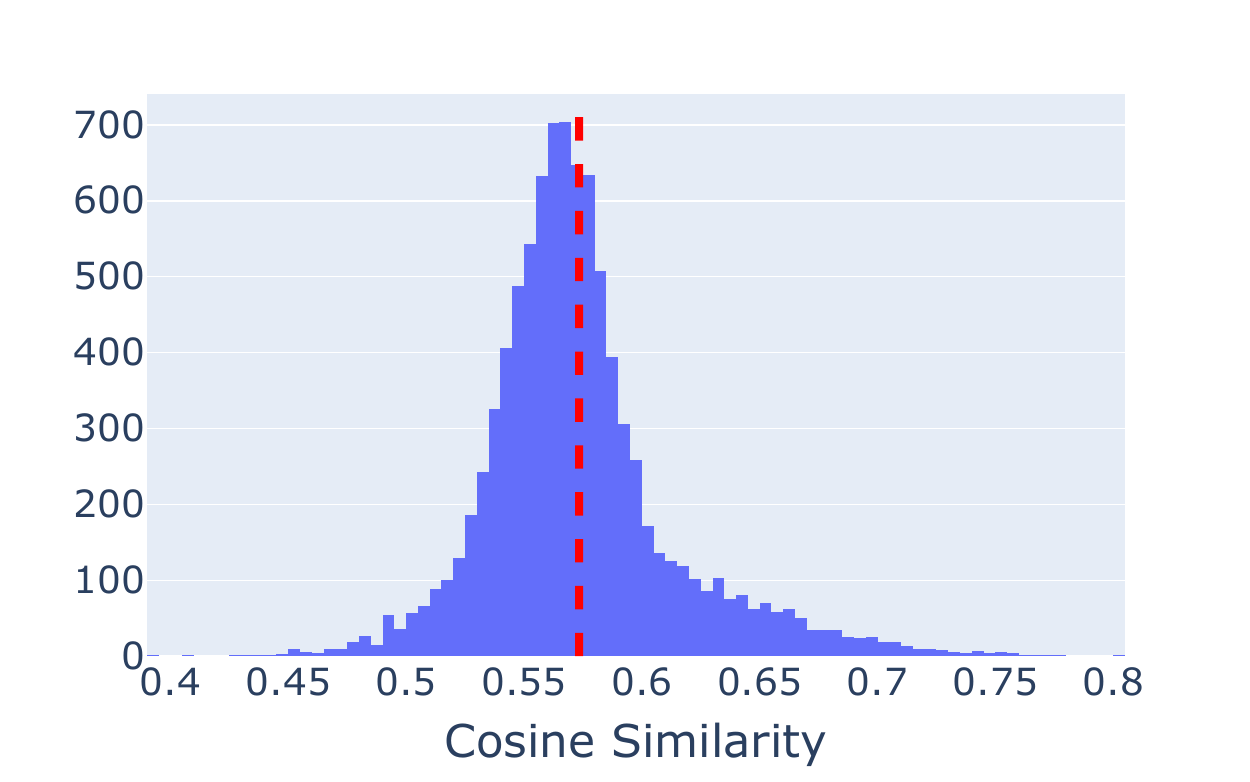}
    \end{minipage}
    \begin{minipage}{0.49\linewidth}
        \includegraphics[width=\linewidth]{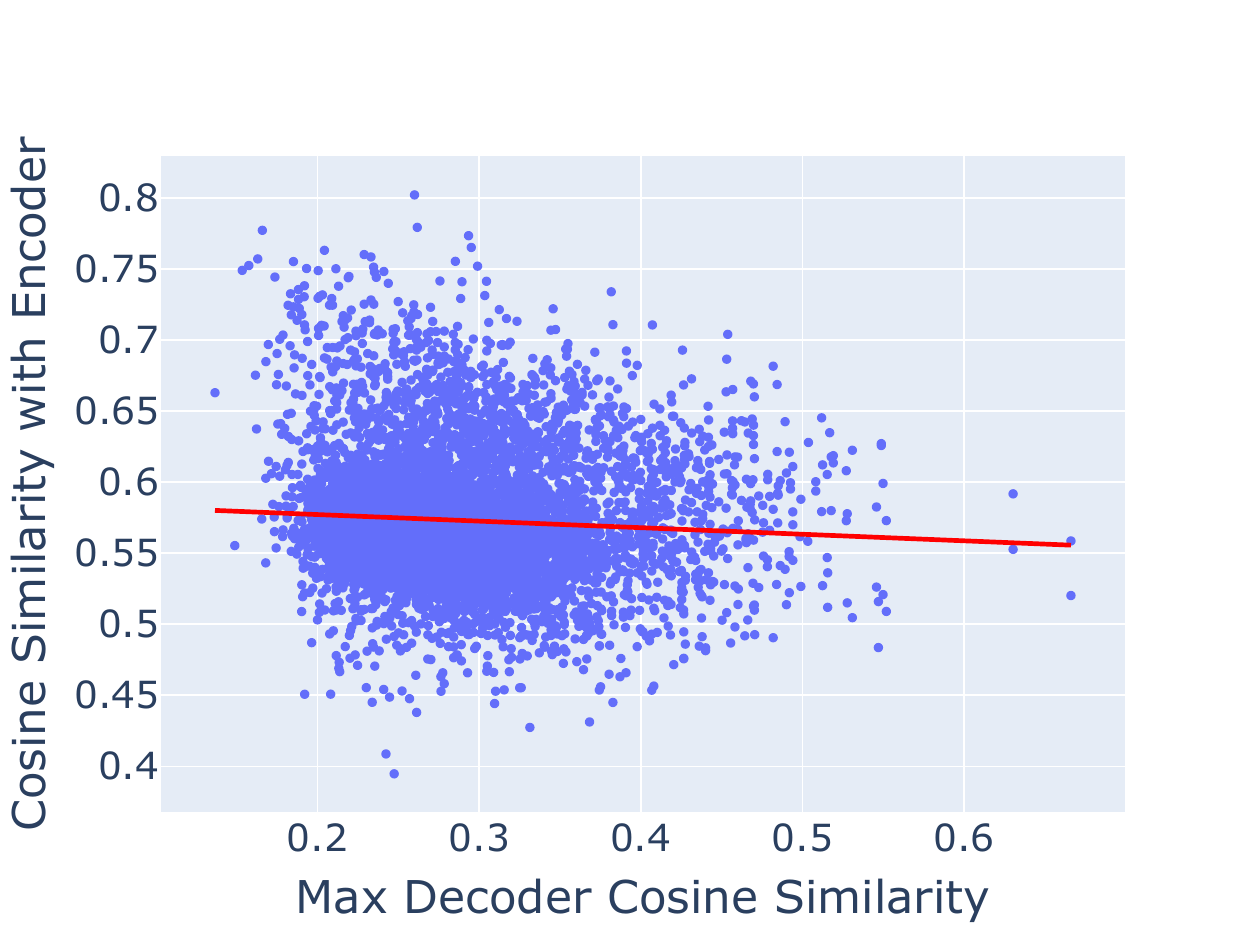}
    \end{minipage}
    \caption{(Left) Cosine similarities between the encoder row and corresponding decoder column for SAE64 (\texttt{cs.LG}). The mean cosine similarity is 0.57, suggesting that encoder and decoder features are rather different, agreeing with \citet{nandaReplication2023}. (Right) We notice a slight negative correlation between a feature's decoder-encoder cosine similarity, and its maximum similarity with other features, possibly suggesting that features that are furthest removed from all other features in embedding space can have more similar corresponding 
    decoders and encoder projections.}
    \label{fig:cosine_sims_encoder_decoder}
\end{figure}

\textbf{Encoder and decoder representations} Figure \ref{fig:cosine_sims_encoder_decoder} reveals an intriguing relationship between feature distinctiveness and the similarity of encoder and decoder representations in our sparse autoencoder. In an ideal scenario with orthogonal features, encoder and decoder vectors would be identical, as the optimal detection direction (encoder) would align perfectly with the representation direction (decoder). This is because orthogonal features can be uniquely identified without interference. However, in our high-dimensional space with more features than dimensions, perfect orthogonality is impossible due to superposition. 

The right panel of Figure \ref{fig:cosine_sims_encoder_decoder} shows a negative correlation between a feature's decoder-encoder cosine similarity and its maximum similarity with other features. Features more orthogonal to others (lower maximum similarity) tend to have more similar encoder and decoder representations. This aligns with intuition: for more isolated features, the encoder's detection direction can closely match the decoder's representation direction. Conversely, features with higher similarity to others require the encoder to adopt a more differentiated detection strategy to minimise interference, resulting in lower encoder-decoder similarity. The left panel, showing a mean cosine similarity of 0.57 between corresponding encoder and decoder vectors, further emphasises this departure from orthogonality. This phenomenon points to the importance of untied weights in sparse autoencoders.

\textbf{Clustering feature vectors}
Motivated by structure in the feature activation graph, we explore whether similar structure can be found in the decoder weight matrix $W$ itself. \cite{gao2024scaling} find 2 such clusters; we reproduce their method across our embeddings and SAEs, permuting the left singular vectors $U$ of $W$ using a one-dimensional UMAP. We also experiment with permuting $U$ and $W$ using reverse Cuthill-McKee. We do not find any meaningful block diagonal structure or clustering in $W$.

\begin{figure}
    \centering
    \begin{minipage}{0.49\textwidth}
        \centering
        \includegraphics[width=\linewidth]{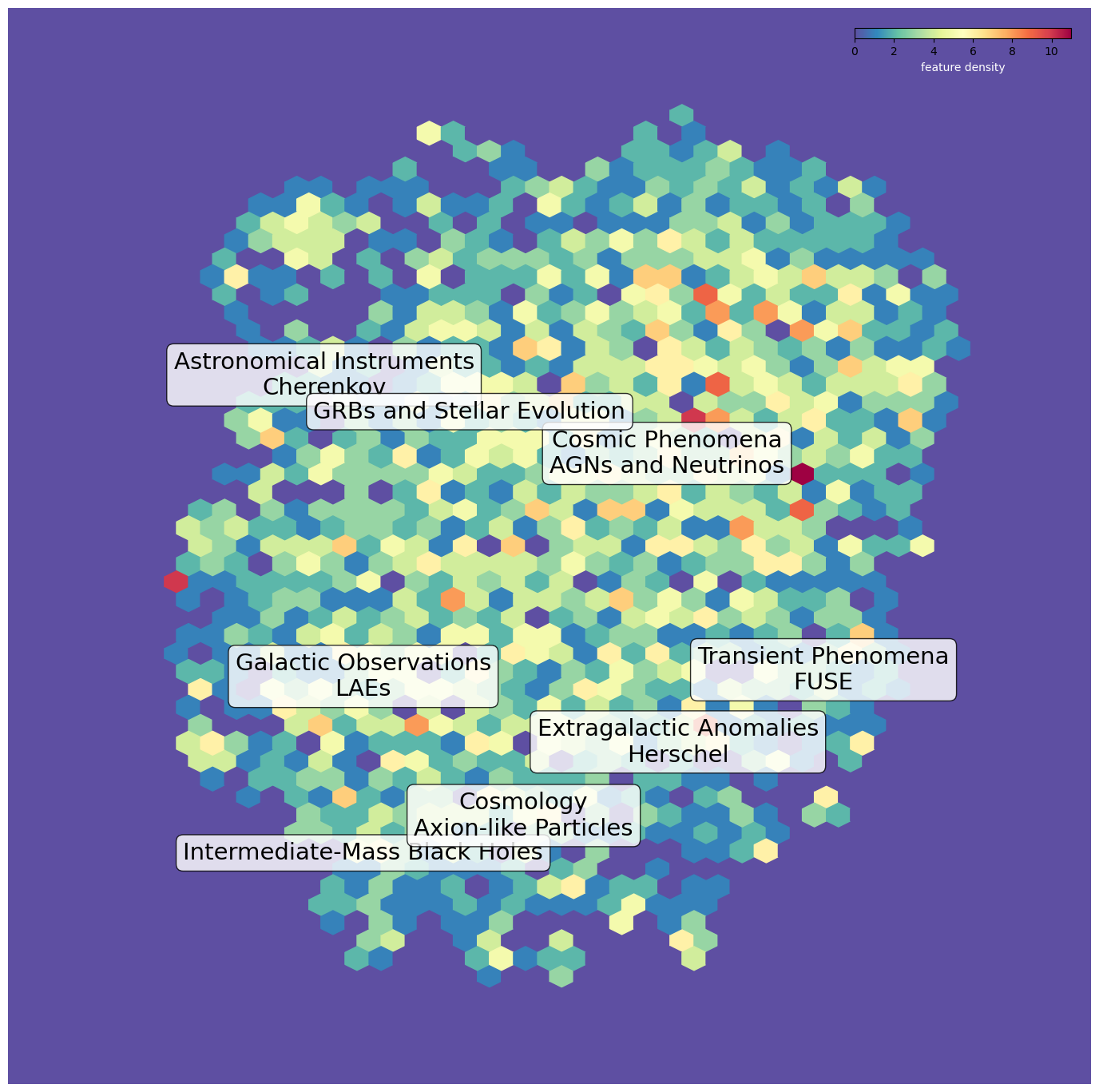}
    \end{minipage}
    \hfill
    \begin{minipage}{0.49\textwidth}
        \centering
        \includegraphics[width=\linewidth]{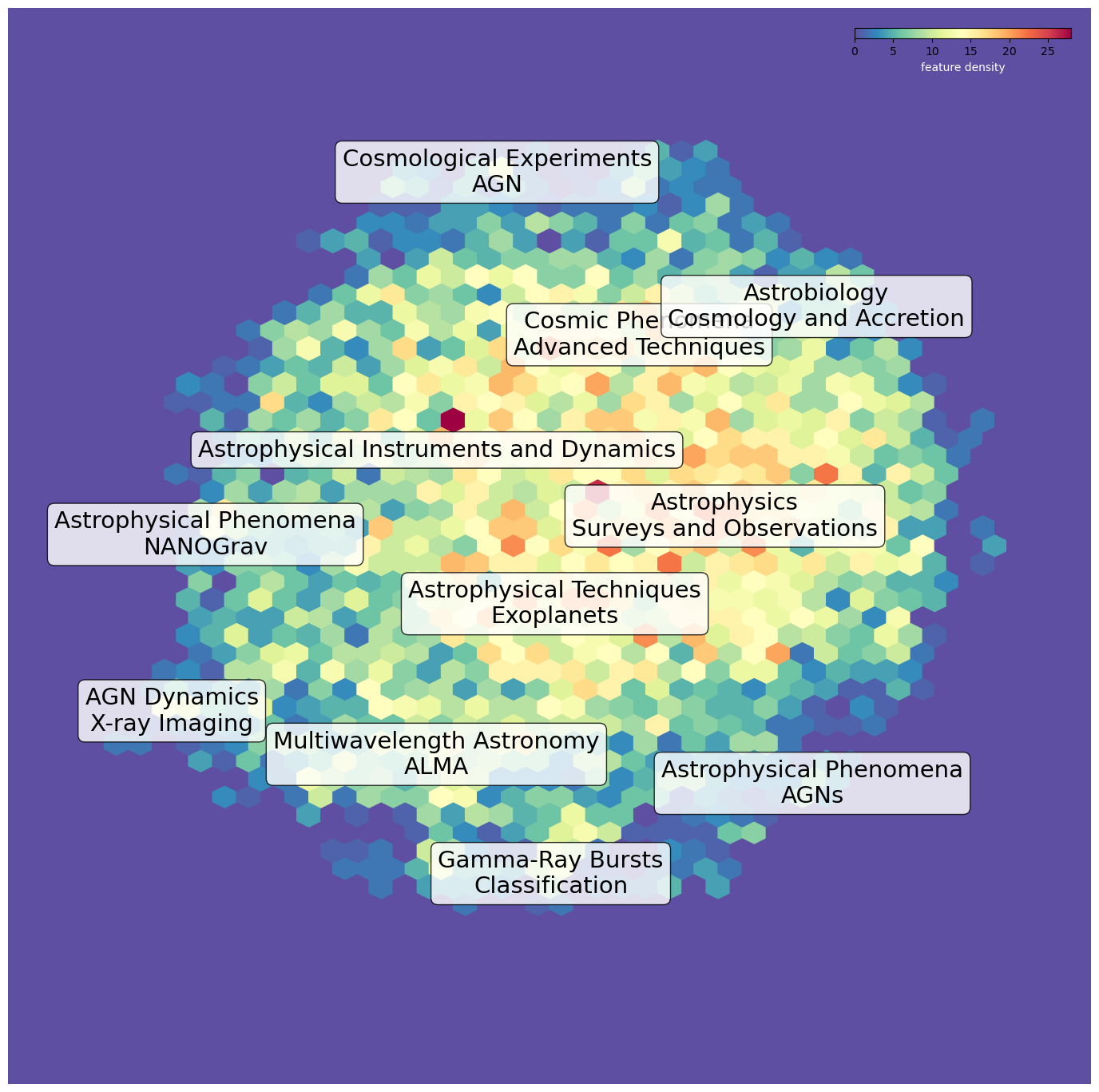}
    \end{minipage}
    \caption{UMAP density plots along with LLM generated labels for SAE16 (left) and SAE64 (right) for the \astroPH{} features.}
    \label{fig:combined-figures}
\end{figure}







\begin{figure}[htpb]
    \centering
    \begin{minipage}{0.48\textwidth}
        \centering
        \includegraphics[width=\linewidth]{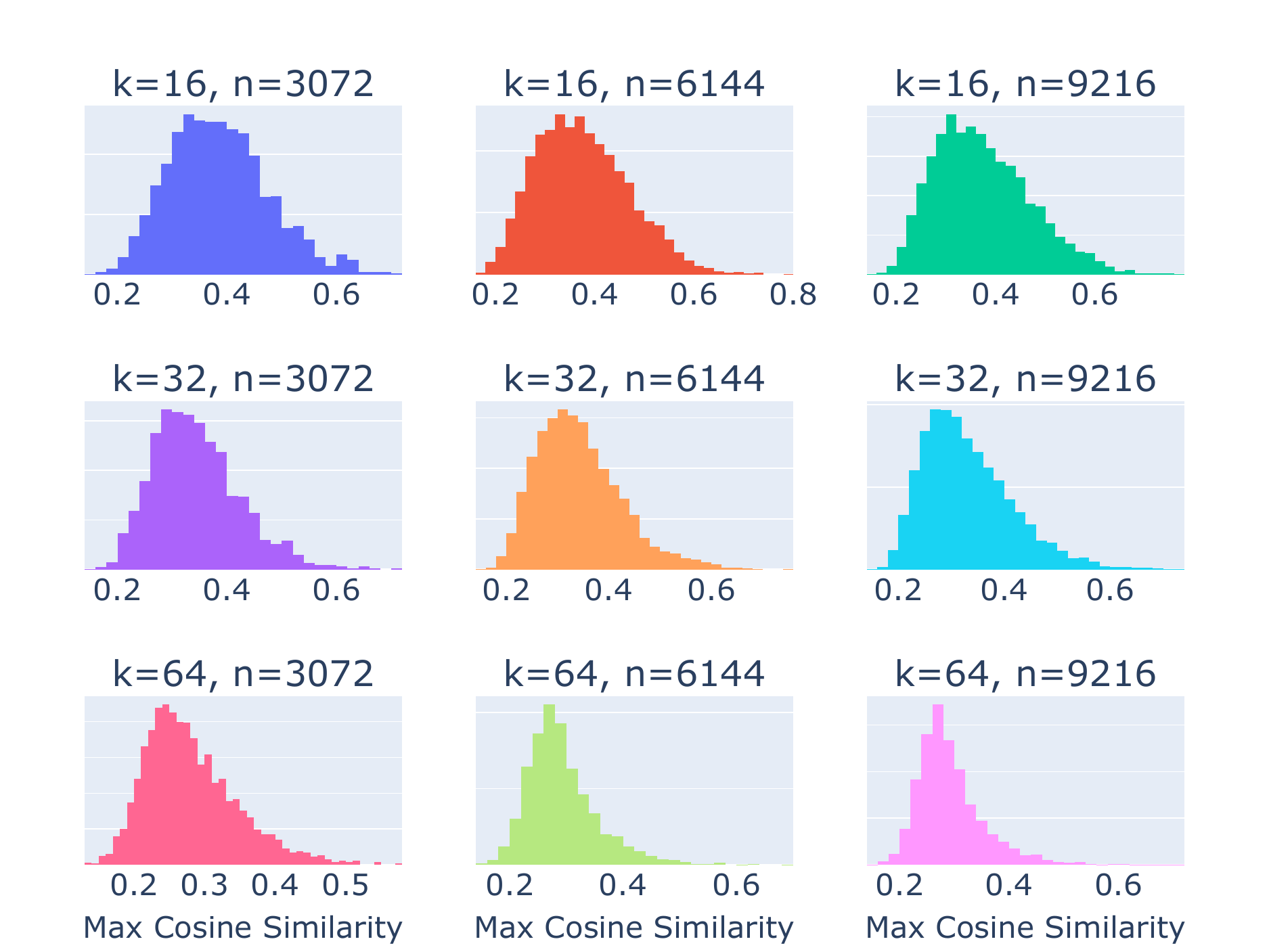}
        \caption{Distribution of maximum cosine similarity between a given feature vector and all other feature vectors, within the same SAE.}
        \label{fig:max_cosine_similarity_histograms_self}
    \end{minipage}\hfill
    \begin{minipage}{0.48\textwidth}
        \centering
        \includegraphics[width=\linewidth]{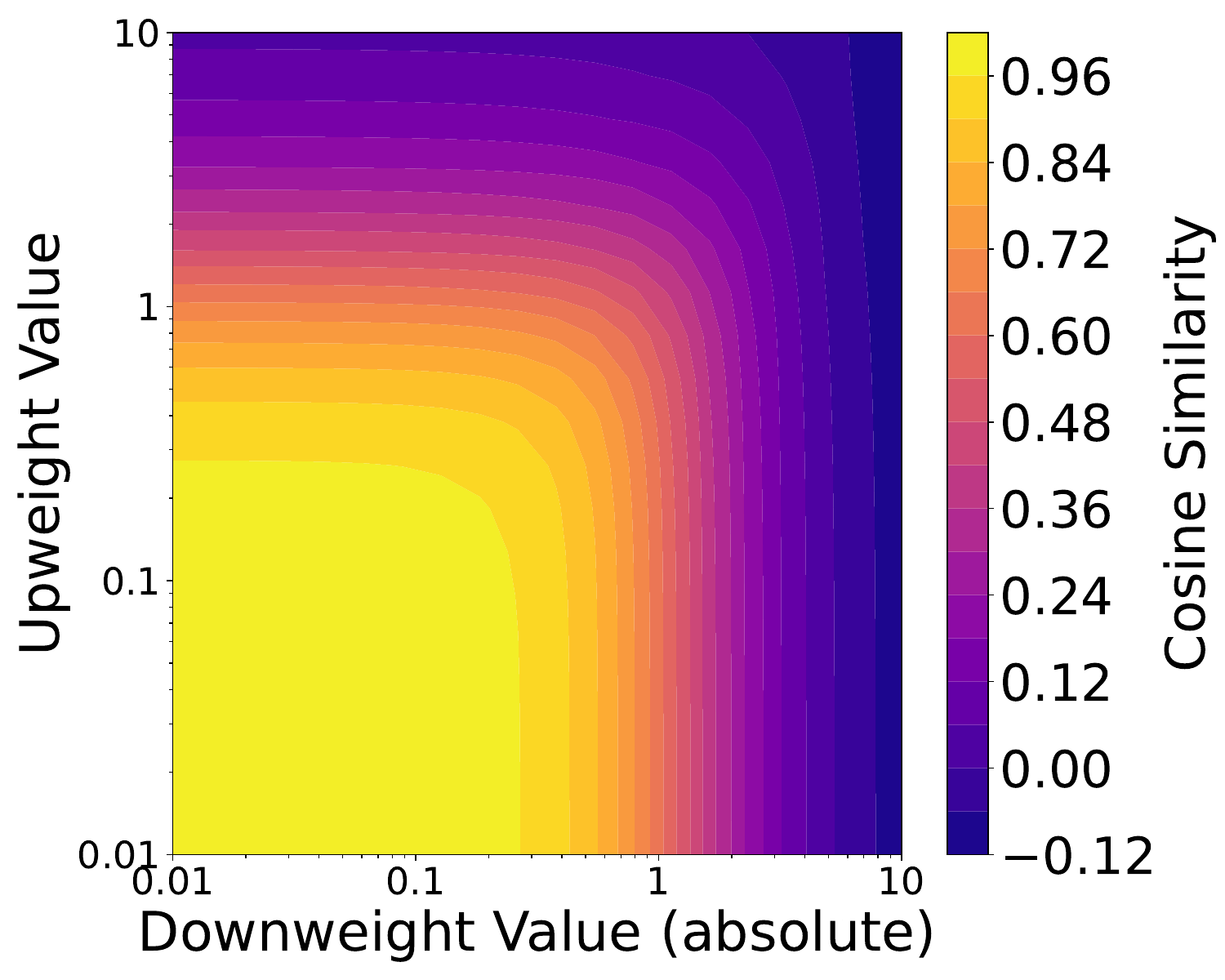}
        \caption{Cosine similarity between the original query embedding and the modified query embedding, with different values of upweighting random zero features and downweighting random active features.}
        \label{fig:cosine_similarity_contour_direct}
    \end{minipage}
\end{figure}

\section{Iterative encoding optimisation}
\label{app:iterative_encoding}

We noted in Section \ref{sec:saerch} that intervening on a feature by up- or down-weighting its hidden representation and then decoding is equivalent to directly adding the scaled feature vector to the final embedding. To demonstrate this equivalence, let's consider an intervention on feature $i$ by an amount $\delta$. The modified hidden representation is $\mathbf{h}' = \mathbf{h} + \delta \mathbf{e}_i$, where $\mathbf{e}_i$ is the $i$-th standard basis vector. Decoding this modified representation gives $\hat{\mathbf{x}}' = W_d\mathbf{h}' = W_d\mathbf{h} + \delta W_d\mathbf{e}_i = \hat{\mathbf{x}} + \delta \mathbf{w}_i$, where $\mathbf{w}_i$ is the $i$-th column of $W_d$. Thus, intervening on the hidden representation and then decoding is equivalent to directly adding the scaled feature vector to the original reconstruction.

We show in Figure \ref{fig:cosine_similarity_contour_direct} how cosine similarity between the original query embedding and the modified query embedding changes as we change the upweighting and downweighting strength for different features. Cosine similarity drops rapidly as soon as upweight or downweight exceeds 0.1.


There is an implicit challenge in SAE-based embedding interventions: the trade-off between steering strength and precision. When directly manipulating feature activations, we observed that strong interventions often led to unintended semantic shifts, activating correlated features and potentially moving the embedding far from the SAE's learned manifold. Our goal is to achieve precise semantic edits that express the desired feature strongly while minimising interference with unrelated features. To this end, we developed an iterative optimisation approach that leverages the SAE's learned feature space to find an optimal balance between these competing objectives.

Let $\mathbf{x} \in \mathbb{R}^d$ be the original embedding, $f_\theta(\cdot)$ the SAE encoder, and $g_\phi(\cdot)$ the SAE decoder. We define a target feature vector $\mathbf{t} \in \mathbb{R}^k$ representing the desired feature activations after intervention, where $k$ is the number of active features in our SAE. The iterative latent optimisation aims to find optimised latents $\mathbf{h}^*$ that satisfy:

$$\mathbf{h}^* = \text{argmin}_{\mathbf{h}'} \left\{\|f_\theta(g_\phi(\mathbf{h}')) - \mathbf{t}\|_2^2\right\}$$

We solve this optimisation problem using gradient descent, starting from the initial latents $\mathbf{h} = f_\theta(\mathbf{x})$ and iteratively updating $\mathbf{h}'$. We use the AdamW optimiser with a cosine annealing learning rate schedule.

To evaluate the effectiveness of this approach, we compare it to a direct intervention method where we simply set the target feature to a specific value in the latent space. For each abstract in our dataset, we embed the abstract using an OpenAI embedding model to obtain $\mathbf{x}$. We then encode the embedding to get initial latents $\mathbf{h} = f_\theta(\mathbf{x})$. We randomly select a target feature $i$ and target value $v$. We then apply both intervention methods: our iterative optimisation of $\mathbf{h}'$ as described above, with $\mathbf{t}_i = v$ and $\mathbf{t}_j = \mathbf{h}_j$ for $j \neq i$, and direct intervention: setting $\mathbf{h}'_i = v$ and $\mathbf{h}'_j = \mathbf{h}_j$ for $j \neq i$.

\begin{figure}[htpb]
    \centering
    \begin{minipage}{0.49\linewidth}
        \includegraphics[width=\linewidth]{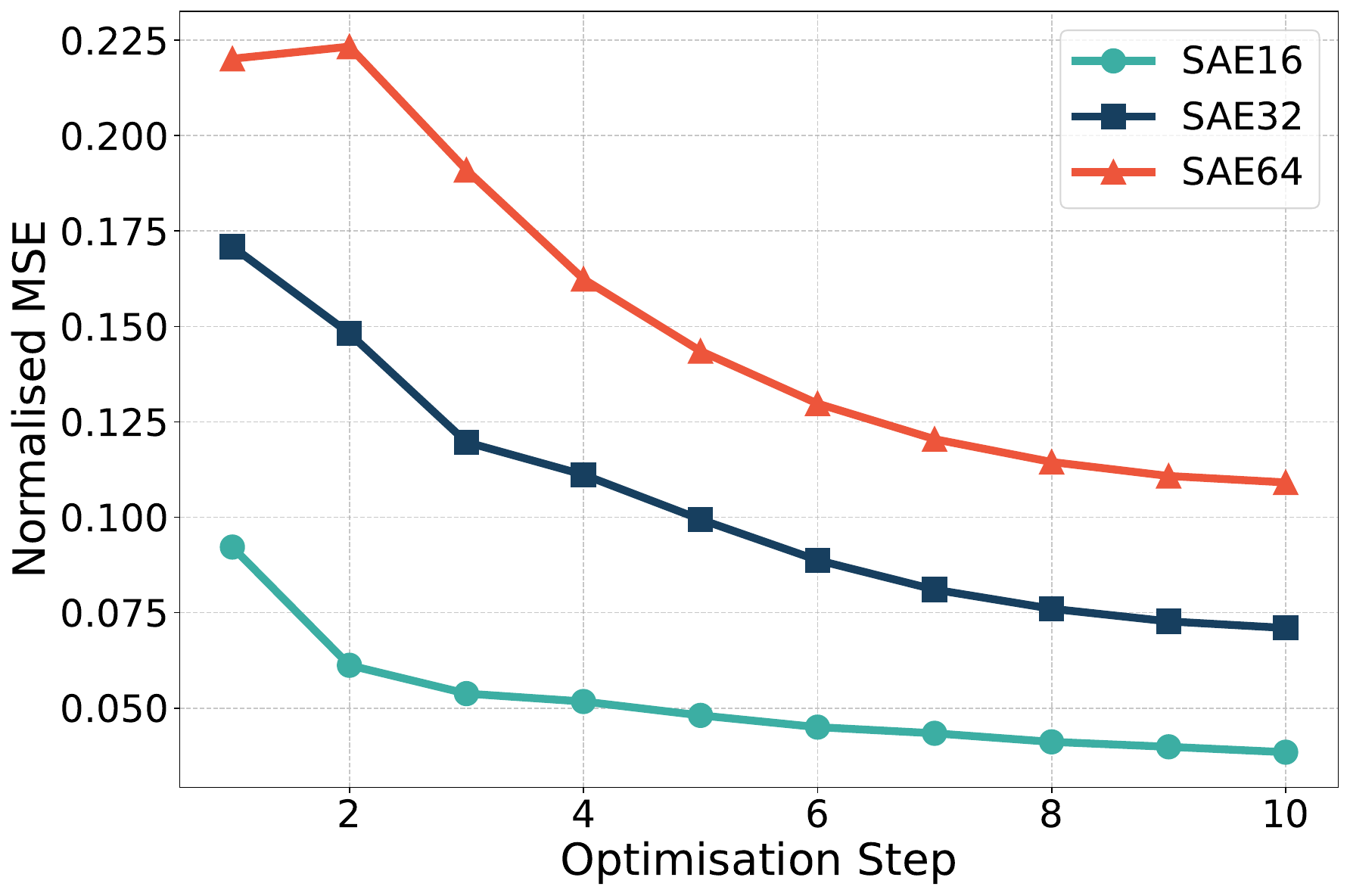}
    \end{minipage}
    \begin{minipage}{0.49\linewidth}
        \includegraphics[width=\linewidth]{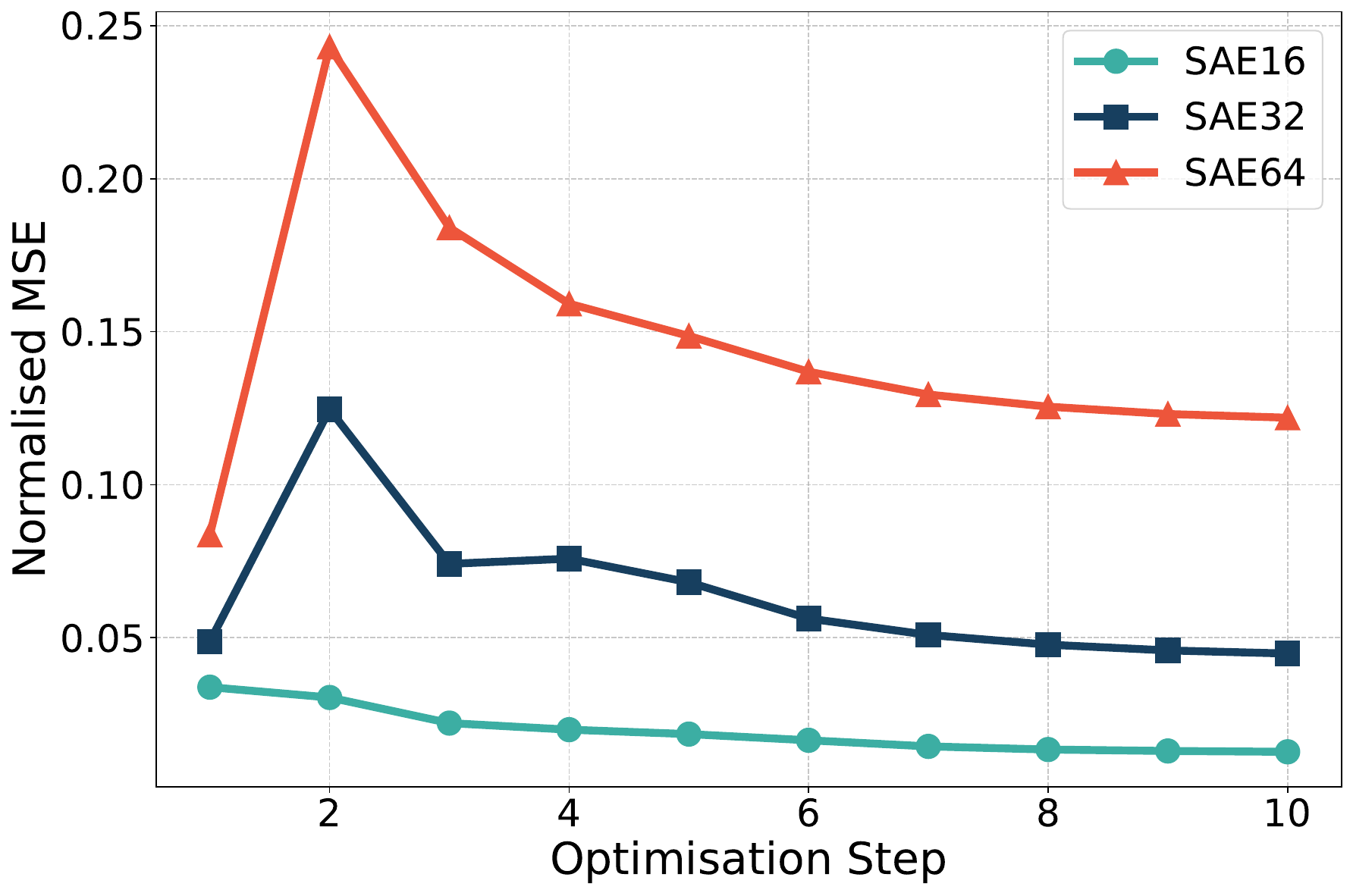}
    \end{minipage}
    \caption{Normalised MSE at each of 10 steps across the iterative latent optimisation process. Left: Setting a random zero feature to active. Right: Setting a random active feature to zero.}
    \label{fig:combined_iterative_optimisation}
\end{figure}

Figure \ref{fig:combined_iterative_optimisation} (left panel) shows the trajectory of normalised MSE during the iterative optimisation process, when setting a random zero feature to active. Similarly, the right panel shows the optimisation when setting a random active feature to zero. Normalised MSE improves in the former case but not the latter.

\end{document}